\PassOptionsToPackage{unicode}{hyperref}
\PassOptionsToPackage{hyphens}{url}
\PassOptionsToPackage{dvipsnames,svgnames,x11names}{xcolor}
\documentclass[
  letterpaper,
  DIV=11,
  numbers=noendperiod]{scrartcl}
\usepackage{xcolor}
\usepackage{amsmath,amssymb}
\usepackage{iftex}
\ifPDFTeX
  \usepackage[T1]{fontenc}
  \usepackage[utf8]{inputenc}
  \usepackage{textcomp} 
\else 
  \usepackage{unicode-math} 
  \defaultfontfeatures{Scale=MatchLowercase}
  \defaultfontfeatures[\rmfamily]{Ligatures=TeX,Scale=1}
\fi
\usepackage{lmodern}
\ifPDFTeX\else
\fi
\IfFileExists{upquote.sty}{\usepackage{upquote}}{}
\IfFileExists{microtype.sty}{
  \usepackage[]{microtype}
  \UseMicrotypeSet[protrusion]{basicmath} 
}{}
\makeatletter
\@ifundefined{KOMAClassName}{
  \IfFileExists{parskip.sty}{%
    \usepackage{parskip}
  }{
    \setlength{\parindent}{0pt}
    \setlength{\parskip}{6pt plus 2pt minus 1pt}}
}{
  \KOMAoptions{parskip=half}}
\makeatother
\makeatletter
\ifx\paragraph\undefined\else
  \let\oldparagraph\paragraph
  \renewcommand{\paragraph}{
    \@ifstar
      \xxxParagraphStar
      \xxxParagraphNoStar
  }
  \newcommand{\xxxParagraphStar}[1]{\oldparagraph*{#1}\mbox{}}
  \newcommand{\xxxParagraphNoStar}[1]{\oldparagraph{#1}\mbox{}}
\fi
\ifx\subparagraph\undefined\else
  \let\oldsubparagraph\subparagraph
  \renewcommand{\subparagraph}{
    \@ifstar
      \xxxSubParagraphStar
      \xxxSubParagraphNoStar
  }
  \newcommand{\xxxSubParagraphStar}[1]{\oldsubparagraph*{#1}\mbox{}}
  \newcommand{\xxxSubParagraphNoStar}[1]{\oldsubparagraph{#1}\mbox{}}
\fi
\makeatother

\usepackage{longtable,booktabs,array}
\usepackage{calc} 
\usepackage{etoolbox}
\makeatletter
\patchcmd\longtable{\par}{\if@noskipsec\mbox{}\fi\par}{}{}
\makeatother
\IfFileExists{footnotehyper.sty}{\usepackage{footnotehyper}}{\usepackage{footnote}}
\makesavenoteenv{longtable}
\usepackage{graphicx}
\makeatletter
\newsavebox\pandoc@box
\newcommand*\pandocbounded[1]{
  \sbox\pandoc@box{#1}%
  \Gscale@div\@tempa{\textheight}{\dimexpr\ht\pandoc@box+\dp\pandoc@box\relax}%
  \Gscale@div\@tempb{\linewidth}{\wd\pandoc@box}%
  \ifdim\@tempb\p@<\@tempa\p@\let\@tempa\@tempb\fi
  \ifdim\@tempa\p@<\p@\scalebox{\@tempa}{\usebox\pandoc@box}%
  \else\usebox{\pandoc@box}%
  \fi%
}
\def\fps@figure{htbp}
\makeatother

\NewDocumentCommand\citeproctext{}{}

\makeatletter
 \let\@cite@ofmt\@firstofone
 \def\@biblabel#1{}
 \def\@cite#1#2{{#1\if@tempswa , #2\fi}}
\makeatother
\newlength{\cslhangindent}
\newlength{\csllabelwidth}
\newenvironment{CSLReferences}[2] 
 {\begin{list}{}{%
  \setlength{\itemindent}{0pt}
  \setlength{\leftmargin}{0pt}
  \setlength{\parsep}{0pt}
  \ifodd #1
   \setlength{\leftmargin}{\cslhangindent}
   \setlength{\itemindent}{-1\cslhangindent}
  \fi
  \setlength{\itemsep}{#2\baselineskip}}}
 {\end{list}}
\usepackage{calc}

\providecommand{\tightlist}{%
  \setlength{\itemsep}{0pt}\setlength{\parskip}{0pt}}

\usepackage{booktabs}
\usepackage{longtable}
\usepackage{array}
\usepackage{multirow}
\usepackage{wrapfig}
\usepackage{float}
\usepackage{colortbl}
\usepackage{pdflscape}
\usepackage{tabu}
\usepackage{threeparttable}
\usepackage{threeparttablex}
\usepackage[normalem]{ulem}
\usepackage{makecell}
\usepackage{xcolor}
\usepackage{ebgaramond}
\usepackage{geometry}
\usepackage{float}
\KOMAoption{captions}{tableheading}
\makeatletter
\@ifpackageloaded{caption}{}{\usepackage{caption}}
\AtBeginDocument{%
\ifdefined\contentsname
  \renewcommand*\contentsname{Table of contents}
\else
  \newcommand\contentsname{Table of contents}
\fi
\ifdefined\listfigurename
  \renewcommand*\listfigurename{List of Figures}
\else
  \newcommand\listfigurename{List of Figures}
\fi
\ifdefined\listtablename
  \renewcommand*\listtablename{List of Tables}
\else
  \newcommand\listtablename{List of Tables}
\fi
\ifdefined\figurename
  \renewcommand*\figurename{Figure}
\else
  \newcommand\figurename{Figure}
\fi
\ifdefined\tablename
  \renewcommand*\tablename{Table}
\else
  \newcommand\tablename{Table}
\fi
}
\@ifpackageloaded{float}{}{\usepackage{float}}
\floatstyle{ruled}
\@ifundefined{c@chapter}{\newfloat{codelisting}{h}{lop}}{\newfloat{codelisting}{h}{lop}[chapter]}
\floatname{codelisting}{Listing}

\makeatother
\makeatletter
\@ifpackageloaded{caption}{}{\usepackage{caption}}
\@ifpackageloaded{subcaption}{}{\usepackage{subcaption}}
\makeatother
\usepackage{bookmark}
\IfFileExists{xurl.sty}{\usepackage{xurl}}{} 
\hypersetup{
  pdftitle={Iterated creation{,} grading{,} and revision of data science projects by language models},
  pdfauthor={Rohan Alexander},
  colorlinks=true,
  linkcolor={blue},
  filecolor={Maroon},
  citecolor={Blue},
  urlcolor={Blue},
  pdfcreator={LaTeX via pandoc}}

\title{Iterated creation, grading, and revision of data science projects
by language models\thanks{\textbf{Keywords:} automated evaluation;
feedback loops; LLM-as-a-judge; reproducibility; data science; agentic
AI. \textbf{Funding information:} I thank Natural Sciences and
Engineering Research Council of Canada (NSERC) Alliance Grant (ALLRP
599949 - 24) and Social Sciences and Humanities Research Council (SSHRC)
Partnership Grant (895-2025-1002) for financial support. I receive free
or heavily subsidized access to frontier models via Claude Code, Codex,
and Antigravity through the academic researcher programs of Anthropic,
OpenAI and Google, respectively. \textbf{AI usage:} I used Claude Code
with Fable 5.1 to help with the coding, including API interactions,
developing the harness, and data pipeline construction, and I used GPT-6
Astra and Gemini Argon as red-team reviewers of the entire project. I
reviewed and evaluated all code and results, and take responsibility for
all aspects. \textbf{Replication materials:}
\url{https://zenodo.org/records/23927861}. \textbf{Thanks:} I thank
Monica Howlett for helpful comments. \textbf{Corresponding author:}
Rohan Alexander:
\href{mailto:rohan.alexander@utoronto.ca}{\nolinkurl{rohan.alexander@utoronto.ca}}.}}
\author{Rohan Alexander}
\date{November 3, 2026}
\begin{document}
\maketitle
\begin{abstract}
We evaluate the extent to which agentic AI models can complete an
end-to-end data science project, and how this improves with feedback. A
`writer' model with specific tools creates a complete data science
project including downloading, cleaning, and analyzing a dataset, and
writing a short paper about it. The project is executed and a `grader'
model evaluates it. The writer can then revise its submission, and this
repeats for a total of three rounds. We consider seven models from five
developers: Anthropic, OpenAI, Google, DeepSeek, and Z.AI. Each model is
both a writer and a grader, so there are 49 writer-grader pairs and we
run each pair three times. Of the 147 initial submissions, 107 run and
render, and the average score increases from 59 per cent to 76 per cent
after the final round of feedback. By writer the extent of the increase
ranges from 3 percentage points for Anthropic's Opus 5.5 to 46
percentage points for Anthropic's Haiku 4.5, and by grader it ranges
from 13 percentage points for Google's Gemini 3.8 Flash to 32 percentage
points for Anthropic's Haiku 4.5. When a model grades its own writing
the score tends to be about 9 percentage points higher than the average
of what the other models give it. About half of the difference that we
see from round to round is within the grader's own noise. Feedback
generally results in the highest marks, but even just having the chance
to revise, or having the additional budget, in terms of time or tool
calls, without any feedback, results in considerable increases for some
of the models but not for others. We find that how much such a loop
helps depends on which model plays which role, but weaker models
generally do not improve stronger models.
\end{abstract}

\newpage

\section{Introduction}\label{introduction}

In many contexts, for many humans, there is great value from feedback,
whether it is from someone better or worse than you. In this paper we
explore if that is also the case for LLMs.\footnote{A large language
  model (LLM) is a neural network trained on a large corpus to predict
  the next token given the tokens before it. LLMs generate output by
  repeatedly sampling one token at a time and appending this to the
  existing output (Alammar and Grootendorst 2024). These capacities can
  be enhanced through tool-use, which means enabling the LLM to, say,
  have commands run in a shell. Grootendorst and Alammar (2026) borrow
  from Russell and Norvig (2020) to say that ``{[}a{]}n agent is
  anything that can be viewed as perceiving its environment through
  sensors and acting upon that environment through actuators''. For
  instance, given a goal and a set of tools, an agentic model may write
  a tool call. A harness, also called a scaffold (Wijk et al. 2024),
  refers to all the software and set-up surrounding the model, for
  instance, to enable the tool call. So in agentic AI, the harness runs
  the call and appends the result to the context, and the model may then
  write another call from that longer context, and so on, until it
  eventually writes the call that ends the task (Yao et al. 2023;
  Grootendorst and Alammar 2026).} Specifically we examine three things.
The first is the quality with which a language model, given specific
agentic capacities, can put together a data science project. The second
is what two rounds of feedback from a `grader' model do to the quality
of that project. The third is how much of this change depends on which
model plays each role.

The data science task that we examine involves writing code to obtain,
clean, test, and analyze data, and writing a paper to communicate the
results. We set up a loop between a `writer' and a `grader' in which a
project is written, graded, revised, graded, revised and graded. First a
`writer' language model creates a complete data science project. When it
does this it works in an agentic way and has access to a shell, a file
writer, a file reader, and a file lister in a container and a budget of
40 minutes and 60 tool calls. It runs its scripts and renders its paper
after it writes, and it submits either when it chooses to or when it
runs out of time or tool calls. After this, the submission is
independently executed in a clean directory, which produces a report of
what ran, what failed, and why. A `grader' model then evaluates the
submission against a rubric with 26 items. When it does this it has
access to the execution report, the files that the writer produced, and
the rubric. In the next round, the writer receives the report and the
grader's score and comment on every rubric item, and can revise its
submission. This happens twice, so each run has three graded
submissions. In our main experiment the writer begins the second and
third round in a new session with a smaller budget of 40 tool calls and
30 minutes, with its own earlier files, the execution report, and the
grader's comment and score for each rubric item, and with nothing else
of the earlier session carried over. We additionally consider two
variations. The first is to keep the three writer rounds but to not have
any feedback and the second is to just have one big unbroken round.

We consider seven models, which are Anthropic's Claude Opus 5.5 and
Claude Haiku 4.5, OpenAI's GPT-6 Sol and GPT-6 Luna, DeepSeek's V4.1
Flash, Google's Gemini 3.8 Flash, and Z.AI's GLM-5.3 Flash. We evaluate
every combination including having a model evaluate its own work. This
gives a total of 49 writer-grader pairs, seven of which involve a model
grading its own writing. Following Cui and Alexander (2026) who advise
repeating configurations independently, we conduct three runs per pair,
for a total of 147 runs. Our choice of models is designed to sample over
the Pareto frontier of capability and cost (Kapoor et al. 2025), along
with a few additions to test specific questions. In particular, we
consider two models from each of Anthropic and OpenAI, one that is more
capable and one that is less capable, to examine the effect of
capability within a developer. From Google we consider an ostensibly
less capable model, but one that does well on benchmarks for its cost.
And we consider open-weight models from DeepSeek and Z.AI. Every model,
apart from DeepSeek, is served through its own developer, via
OpenRouter, at default settings.

Of the 147 initial submissions, we find that 107 run end to end and
render, and after the final revision this increases to 116. The average
score increases from 59 per cent to 76 per cent after the final
revision. About a third of that increase is because fewer submissions
receive zero, but those tend to be concentrated in a few models and if
we remove those submissions, then the increase is from 70 per cent to 80
per cent. The increase depends on both the writer and the grader. For
each of the writers it ranges from 3 percentage points with Opus 5.5 as
the writer to 46 percentage points with Haiku 4.5 as the writer,
although a considerable part of that is how much room for improvement
there was initially. Over each of the graders other than Haiku 4.5,
which gives many submissions 0, it ranges from 13 percentage points with
Gemini 3.8 Flash as the grader to 22 percentage points with GPT-6 Luna
as the grader. Most of the increase comes from the first revision, with
an average increase of 13 percentage points compared with an additional
5 percentage points after the second revision. A model grading its own
final submission scores it about 9 percentage points higher compared
with the other graders grading that same submission. Revising without
any feedback, but with the same additional budget of time and tool calls
in the second and third rounds, gets writers other than Haiku 4.5 some
of the way there. Specifically, on average, to 74 per cent compared with
78 per cent with feedback, when marked by the other graders. And just
having that additional budget of time and tool calls initially, but
without any rounds of revision, gets four of the seven writers to within
three points of their final submission with feedback. That all said,
revising after feedback usually results in the highest scores.

A few other findings are worth highlighting. Firstly, when we examine
the consistency of the marking we find that the same grader scoring the
same final submission three times gives scores that differ by almost 4
percentage points on average. Secondly, the writer does tend to act on
the feedback. Of the 3,564 rubric items that were below the maximum for
that rubric item, the file that the rubric item relates to changed in 76
per cent of cases. Finally, despite doing well overall on this task, the
models name themselves as a model from a different developer in 44 per
cent of final submissions, highlighting the brittleness of the
accomplishment. Nearly all of this is because of models from Google,
DeepSeek, and Z.AI saying they were from Anthropic.

The extent to which AI models can do data science has been explored in
various settings. Many earlier benchmarks of data science agents gave
the model a dataset and a task with a specific outcome, such as a
numerical answer, a set of analysis decisions, or a Kaggle leaderboard
score (Z. Chen et al. 2025; Chan et al. 2025; Jing et al. 2025), but
real-world data science tends to be less clear-cut. Previous work has
found that LLMs can produce every part of such a project, including
grading it (Hughes and Alexander 2026), but their writer did not have
substantial tool-use and their grader did not execute the scripts. More
recent work tends to focus on agentic approaches. For instance, Rahman
et al. (2026) benchmark agents on 275 data science workflow tasks,
finding that the best agent completes 57 per cent of them, but the tasks
have specified goals and no feedback mechanism. Even agents that run
their analyses end-to-end still lack reliable statistical reasoning
(Miao et al. 2026). And frontier models were found to fabricate data,
when presented with situations where they should actually acknowledge
failure (Yang et al. 2026). In terms of open-ended research, Edwards et
al. (2026) find that some models can sometimes implement extensions to
published research papers, while Kim et al. (2026) find less promising
agentic qualities in terms of writing follow-up papers on open problems,
including finding that models claim to have done work that they did not.

We use LLM-as-a-judge, which has been found to have lower agreement with
experts on tasks that need domain knowledge (Szymanski et al. 2025), and
to give scores that change with the order of the rubric's score
descriptions and with the score attached to a reference answer (Li et
al. 2026). The issue of self-bias in particular is well documented
(Panickssery et al. 2024; W.-L. Chen et al. 2025; Roytburg et al. 2026).
That said, AI models have been found to give useful feedback. For
instance, a randomized study at ICLR 2025 found that when reviewers were
given AI feedback on their reviews, they updated their reviews about a
quarter of the time, and that revised reviews tended to be preferred
(Thakkar et al. 2025), and Bianchi et al. (2026) ran a conference whose
papers were primarily written and reviewed by AI. In the social sciences
King and Worthington (2026) find that AI-generated reviews of articles
tended to be better than reviews written by paid human reviewers.

Beyond data science, the type of loop that we examine in this paper is
one way in which AI is expected to make automated contributions to
science (Swanson et al. 2025; Lange et al. 2026; Liu et al. 2026). That
is, an AI model produces some output, which is judged by an AI model,
and then the original model revises its output. Expectations for the
improvements due to such loops turn on whether the AI judgment has a
significant effect. Our set-up allows us to explore aspects of that
larger question. For instance whether a weaker model can usefully advise
a stronger one, the extent of stochasticity in outcomes, and whether it
is the feedback or the opportunity to keep going that matters. Our
approach could be widely adapted in any area where knowledge work occurs
and can be evaluated.

\section{Data and methods}\label{sec-data}

\subsection{Task}\label{task}

The task that we evaluate is based on a broad assignment given to
third-year undergraduate statistics students. Students are expected to
develop a research question that is of interest to them and then obtain
or create a relevant dataset and put together a paper that answers it.
They use \texttt{R} or \texttt{Python} to download, clean, and analyze
the dataset, and then use Quarto to write a 5-20 page paper including
graphs, tables, and references, that communicates their work and
findings. The expectation is that all of the analysis is reproducible
and students submit a link to a GitHub repository with all their files.
Students are provided with a template repository, past examples, and the
rubric by which they will be graded. The rubric has 28 items and a total
of 112 points. Further details about the rubric are available in
Appendix~\ref{sec-rubric-summary}.

We present essentially the same set-up and broad task to the models
although two items, hand-drawn sketches and a commit history, are
removed from the rubric that the models are evaluated against, leaving
26 items and a total of 108 points. While the task itself is essentially
the same, the broader context of the task is entirely different. The
students are completing the task as part of their learning. Before
taking the class some students may have never written code before, and
others may be new to writing a paper. Students almost always have
competing priorities and other commitments, but they can work as much as
they like on this project, typically over a one-to-two-week period, and
they get one round of feedback. In contrast, the models that we consider
have been deliberately trained to do well at generating code to achieve
some outcome. We provide them with a budget of 60 tool calls and a
maximum of 40 minutes for the first round, and an additional 40 tool
calls and 30 minutes for each of the subsequent two rounds. This is the
same for the variation without feedback, while in the variation where
the writer receives this all initially they get 140 tool calls and 100
minutes.

\subsection{Models}\label{models}

The seven models that we consider are Anthropic's Claude Opus 5.5 and
Claude Haiku 4.5, OpenAI's GPT-6 Sol and GPT-6 Luna, DeepSeek's V4.1
Flash, Google's Gemini 3.8 Flash, and Z.AI's GLM-5.3 Flash. We use all
the models as both a writer and a grader. We were not able to evaluate
Anthropic's Fable, OpenAI's Astra, or Google's Argon models because of
cost constraints. Details of our total costs are available in
Appendix~\ref{sec-costs}.

We queried each model, apart from DeepSeek V4.1 Flash, using OpenRouter.
We pinned our requests to the developer's own host and disabled fallback
to other hosts. We also checked the responses for consistency between
the actual developer's API and OpenRouter
(Appendix~\ref{sec-api-check}). For V4.1 Flash we used the DeepSeek API
directly because OpenRouter's DeepSeek-served endpoint was not available
to us when we conducted our runs between 28 and 30 September 2026.

\subsection{Writer-grader interaction}\label{writer-grader-interaction}

\begin{figure}

\centering{

\pandocbounded{\includegraphics[keepaspectratio]{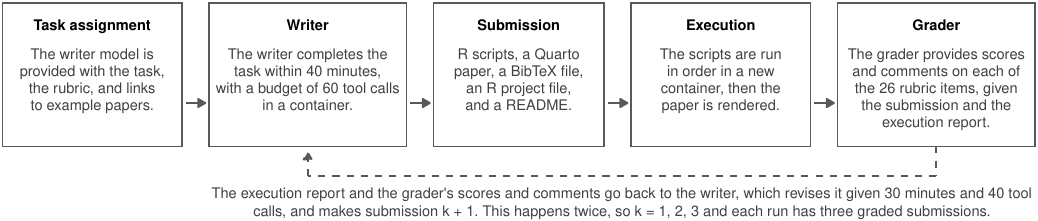}}

}

\caption{\label{fig-loop}In each of three rounds the writer creates a
submission, the submission is executed in a new container, the grader
evaluates it with the files and an execution report, and the report and
the grader's scores and comments go back to the writer.}

\end{figure}

We consider three rounds of interaction between the writer and the
grader (Figure~\ref{fig-loop}). Each of the seven models acts as both a
writer and grader, which gives 49 pairs, and in seven of these a model
grades its own writing. Each pair is independently run three times to
provide some allowance for the stochasticity of the outputs. The writer
creates a new first submission in every run, so the 21 first submissions
of each writer are all different, and the writer does not know which
model will grade it.

The first step is that the writer creates a project of eleven files in a
working directory. This includes \texttt{R} scripts for simulating,
downloading, cleaning, testing, exploring, and analyzing the data, a
Quarto file containing the paper, a BibTeX file for citations, an
\texttt{R} project file, and a README. The writer has access to five
tools. These are to run a shell command, write a file, read a file, list
the files, and to end the session. A session ends when the writer ends
it, when it has used its tool calls or its time, or when it replies
without calling a tool, and the submission is the files in the directory
at that point, but does not include the data that are used or the
rendered output because everything is expected to be able to run and
render in a new environment. The prompts that we use are available in
Appendix~\ref{sec-prompts}.

The submission is then evaluated in a new container. The \texttt{R}
scripts are run in filename order, and the paper is rendered with
Quarto. These run inside a container that has R, Python, Quarto, and a
wide range of packages. The container has network access, which is
typically needed to download data. A neutral executor evaluates whether
the download and cleaning scripts exited without error and wrote at
least one data file, whether the test and model scripts exited without
error, and whether the Quarto file produced a PDF. A submission is
considered to execute if the scripts run without errors and the paper
renders. The executor produces a report that lists the five outcomes,
each script's exit status and running time, and the error lines for any
failed scripts or render.

The grader then receives the rubric, all the files that the writer
created, and the execution report and returns a score and a comment for
each rubric item. The rubric is strict and some breaches, such as using
simulated rather than real data, even result in getting zero overall. If
a grade cannot be read, or there is a rubric item that does not have a
mark, then the grader is given opportunities to address this, and if
there is still not a mark for every rubric item, then the round's grade
is considered unavailable. The score of a submission is worked out as a
percentage, rounded to the nearest whole number.

After the first round is completed the writer is given back the files
that it submitted, as well as the execution report, and the grader's
comment and score for every rubric item. The writer revises its
submission with the same tools, and the new submission is executed and
graded in the same way. We consider three rounds in total because while
having models revise their own output has been found to be useful in
some settings, the increases are typically found to diminish with each
additional round (Chen et al. 2024; Madaan et al. 2023).

We consider two main variations. In the first, the writer has everything
it has in a run with feedback except what comes back from the grader.
That is, it has the same 40 tool calls and 30 minutes, it starts from
the same files, and it can run its own scripts and render its own paper,
so it can, in theory, find the errors that the execution report would
have told it about. Each of the 146 runs whose three submissions were
graded is continued in this way from its first submission, which gives
292 further submissions. In the second, to evaluate the effect of work
coming in three pieces, we also had each writer run three single
sessions with the whole allowance of the three rounds, 140 tool calls
and 100 minutes, told the budget at the start, making one submission in
each with no revision, and had every grader mark each of them
(Appendix~\ref{sec-no-feedback}).

We also consider two additional evaluations. Firstly, to compare the
graders on the same submissions, we also had each of the seven graders
grade the first and final submission of every run, and had the grader of
each run grade them once more. Secondly, in Appendix~\ref{sec-grader-ta}
we compare the marks given by each grader with the marks that human
teaching assistants gave submissions completed by students.

\section{Results}\label{sec-results}

\subsection{Execution}\label{sec-res-reproducibility}

In terms of whether the scripts run or not, we find that 107 of the 147
first-round submissions run end to end and render
(Table~\ref{tbl-repro-writer}). Across all three rounds, 111 of the 441
submissions do not execute, and 17 runs do not execute in any of the
three rounds, 16 of them written by Z.AI's GLM-5.3 Flash. Of the
submissions that do not execute, the paper does not render in 86 per
cent of cases and a script fails in 61 per cent of cases.

\begin{table}

\caption{\label{tbl-repro-writer}The number of submissions that execute
in the first round, in the final round, and for all three rounds, by
writer. For the 21 submissions it made in the first round, for each
model, we show the average number of minutes the writer took, the
average number of tool calls, and how the session ended (the writer
ended it, it used all 60 tool calls, it reached the 40-minute limit, or
it replied without calling a tool). One third-round submission of GPT-6
Luna was too large for its grader to read and does not have a score but
is included here because it executes.}

\centering{

\centering\begingroup\fontsize{8}{10}\selectfont

\resizebox{\ifdim\width>\linewidth\linewidth\else\width\fi}{!}{
\begin{tabular}{lrrrrrrrrr}
\toprule
\multicolumn{1}{c}{ } & \multicolumn{3}{c}{Submissions that execute} & \multicolumn{2}{c}{First submission} & \multicolumn{4}{c}{How the session ended} \\
\cmidrule(l{3pt}r{3pt}){2-4} \cmidrule(l{3pt}r{3pt}){5-6} \cmidrule(l{3pt}r{3pt}){7-10}
Writer & First round & Final round & Any round & Minutes & Tool calls & Writer ended & Tool limit & Time limit & No tools\\
\midrule
Opus 5.5 & 21 of 21 & 21 of 21 & 63 of 63 & 31.1 & 43 & 12 & 1 & 0 & 8\\
Haiku 4.5 & 18 of 21 & 18 of 21 & 49 of 63 & 9.7 & 58 & 15 & 6 & 0 & 0\\
GPT-6 Sol & 21 of 21 & 21 of 21 & 63 of 63 & 17.3 & 56 & 21 & 0 & 0 & 0\\
GPT-6 Luna & 18 of 21 & 19 of 21 & 54 of 63 & 12.7 & 58 & 3 & 17 & 1 & 0\\
DeepSeek V4.1 Flash & 11 of 21 & 20 of 21 & 47 of 63 & 17.5 & 58 & 1 & 18 & 2 & 0\\
Gemini 3.8 Flash & 17 of 21 & 12 of 21 & 45 of 63 & 7.2 & 58 & 11 & 9 & 0 & 1\\
GLM-5.3 Flash & 1 of 21 & 5 of 21 & 9 of 63 & 36.1 & 50 & 1 & 11 & 7 & 2\\
\bottomrule
\end{tabular}}
\endgroup{}

}

\end{table}%

Most of the models are able to create submissions that execute. The
notable exception is GLM-5.3 Flash for which only 1 of its 21 initial
submissions execute. That model also tends to be the slowest, taking 36
minutes on average, compared with the average of the other models being
16 minutes. It reaches the first-round 40-minute time limit in 7 of its
21 submissions. DeepSeek V4.1 Flash also does less well here with only
11 of the 21 initial submissions executing.

Of all the first submissions 73 per cent run and render, and of all the
final submissions 79 per cent do after the two rounds of feedback. This
is similar to the 80 per cent that run and render after two rounds of
revision without feedback (Appendix~\ref{sec-no-feedback}). But these
similar averages hide different underlying processes. Feedback does, of
course, result in submissions running and rendering that previously did
not, but feedback can also be bad and cause submissions that did run and
render to stop. In the variation without feedback, although we find that
fewer broken submissions are fixed, it is also the case that fewer
working submissions are broken.

\subsection{Overall performance by writer}\label{sec-res-runs}

\begin{figure}[p]

\centering{

\pandocbounded{\includegraphics[keepaspectratio]{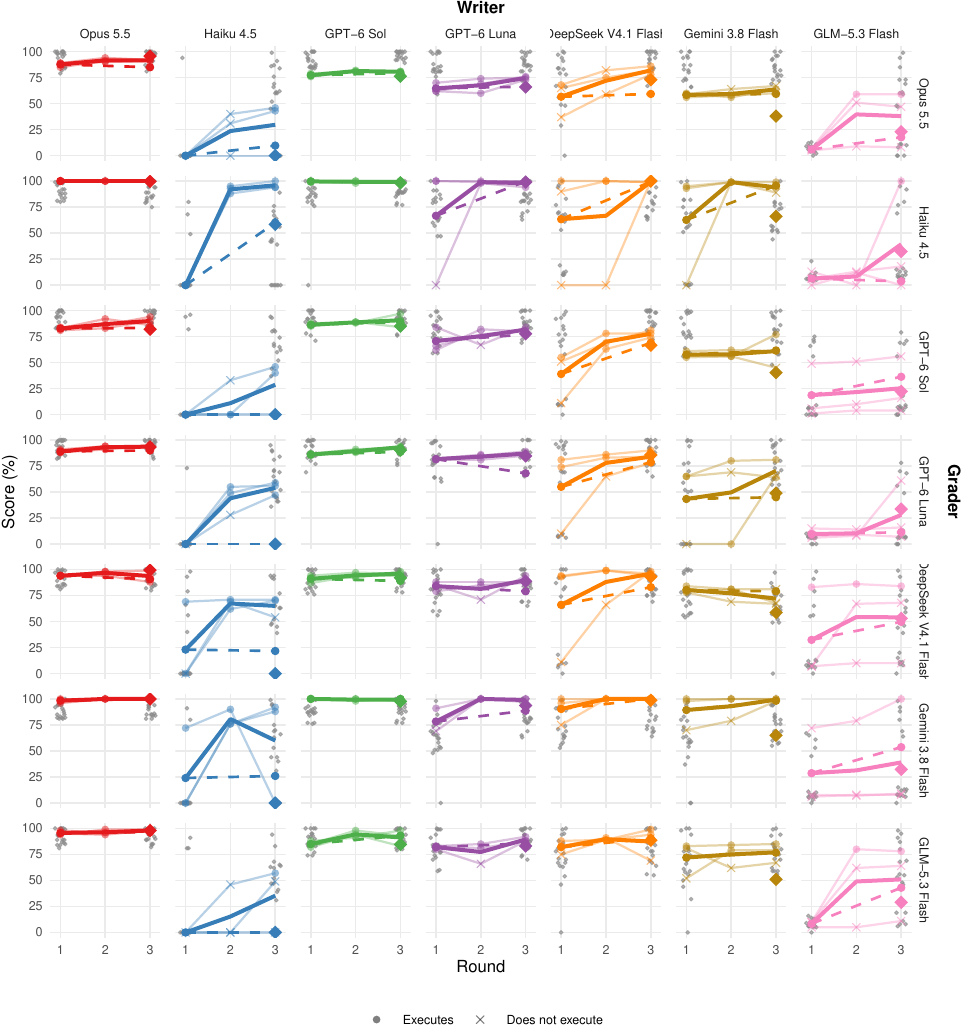}}

}

\caption{\label{fig-runs}Percentage score, by round, for each of three
runs, with one panel for each writer-grader pair. Each run is a thin
line and the thick line is the average of the pair's three runs, as
scored by that pair's grader. A cross indicates a submission that does
not execute. The small grey points at rounds 1 and 3 are the scores that
the six graders other than the pair's grader give the same submissions.
The dashed line shows how the average of three scores changes after two
rounds without feedback. The diamond at round 3 only is the average of
three scores after one session with the whole allowance.}

\end{figure}

Performance can be evaluated based on the score that a submission
received in the first round, compared with the score that it eventually
received at the end of the final round (Figure~\ref{fig-runs}). On
average, and across all the runs, the score increases from 59 per cent
in the first round to 76 per cent in the final round. Most of this comes
from changes made after the first round of feedback, which increases the
average score by 13 percentage points. In that second round, 105 of the
147 runs increase, 24 stay the same, and 18 decrease, compared with an
average increase of 5 percentage points in the third round, with 77
increasing, 34 staying the same, and 35 decreasing. Even after the third
round, the scores from three independent runs for each writer-grader
pair have an average difference of 21 percentage points, although this
is mostly because of a small number of pairs with substantial
differences. When GLM-5.3 Flash is removed as writer and grader it is 11
percentage points, and when Haiku 4.5 is also removed it is 7 percentage
points. Of the 147 runs, 30 runs score below their first submission in
some round and 16 runs end below it. Part of the increase across the
rounds is because of the rules that give 0 overall. In the first round
24 submissions receive 0 because of at least one of them, in the second
it is 8 and in the third it is only 5.

However, some nuance is warranted. A writer that did well initially does
not have much scope to improve regardless of the quality of the feedback
that it receives from the grader. But also a writer may not improve
considerably if it does not take on board useful feedback. Similarly, a
grader may not have much room to suggest improvements if the initial
submission is of high quality. But also a grader may not result in
substantive improvements if the quality of its suggestions is not
sufficiently high, or if they are not implemented. There is also
stochasticity in terms of the outputs.

As a writer, Anthropic's Opus 5.5 has an average score of 92 per cent in
the first round, which is the highest of any writer, and 95 per cent in
the final round, an increase of 3 percentage points. The less capable
model from Anthropic that we consider, Haiku 4.5, has an average score
of 7 per cent in the first round, which is the lowest of any writer, and
increases, on average, by 46 percentage points, which is the largest
increase of any writer. When all seven graders score the final
submissions, those of Haiku 4.5 average 51 per cent, and those of the
other writers average between 37 and 93 per cent. As a grader, Opus 5.5
gives first-round submissions an average of 50 per cent, and the
submissions it grades increase by 16 percentage points, with 20 of the
21 runs that it grades ending above their first-round score. Haiku 4.5
gives first-round submissions an average of 57 per cent, gives 100 per
cent to 41 per cent of the submissions it grades, and the submissions it
grades increase by 32 percentage points, which is the largest increase
of any grader. However, much of that increase is because Haiku 4.5 gives
0 per cent to seven of the 21 first-round submissions that it grades.
Three of the seven are its own, and most of the other graders also give
those 0 per cent. The other four are from other writers, and none of the
other six graders gives any of them 0 per cent. Amusingly, the
first-round submissions of Haiku 4.5 receive an average of 0 per cent
from Opus 5.5, 0 per cent from GPT-6 Sol, and 0 per cent from Haiku 4.5
itself, and so it at least recognizes that its own submissions are poor.

As a writer, OpenAI's GPT-6 Sol has an average score of 89 per cent in
the first round, which is the second highest of any writer, and
increases by 4 percentage points after the final round. The other model
from OpenAI that we evaluate, GPT-6 Luna, has an average score of 75 per
cent in the first round and increases by 13 percentage points, and none
of its 21 runs end below their first-round score, which is the fewest of
any writer. When all seven graders score the final submissions, those of
GPT-6 Sol average 92 per cent, which is the second highest after Opus
5.5, and those of GPT-6 Luna average 86 per cent. As graders, GPT-6 Sol
and GPT-6 Luna give first-round submissions an average of 51 per cent
and 52 per cent, which along with 50 per cent from Opus 5.5 are the
lowest of any grader, and neither gives any submission 100 per cent. The
submissions that GPT-6 Sol and GPT-6 Luna grade increase by 14 and 22
percentage points. With GPT-6 Sol as the grader two of its 21 runs end
below their first-round score, and with GPT-6 Luna two do, compared with
four with DeepSeek V4.1 Flash, which is the most of any grader.

As a writer, Google's Gemini 3.8 Flash has an average score of 66 per
cent in the first round and increases by 11 percentage points. When all
seven graders score the final submissions, those of Gemini 3.8 Flash
average 74 per cent. As a grader, Gemini 3.8 Flash gives 100 per cent to
51 per cent of the submissions it grades, and to 6 of its 21 first-round
submissions. No other grader apart from Haiku 4.5 gives 100 per cent to
more than one first-round submission. A first-round submission that
receives 100 per cent does not give the writer much to act on and so the
run usually stays at 100 per cent through subsequent rounds. The
submissions that Gemini 3.8 Flash grades increase by 13 percentage
points, partly because of this. It is interesting that despite giving
100 per cent to so many submissions from so many models, the one that it
does not give it to is the less capable model from Anthropic, Haiku 4.5.

As a writer, DeepSeek V4.1 Flash has an average score of 65 per cent in
the first round and increases by 25 percentage points. When all seven
graders score the final submissions, those of DeepSeek V4.1 Flash
average 87 per cent. As a grader, DeepSeek V4.1 Flash gives first-round
submissions an average of 67 per cent and never gives 100 per cent. The
submissions that it grades increase by 14 percentage points, and four of
its 21 runs end below their first-round score.

As a writer, GLM-5.3 Flash has an average score of 16 per cent in the
first round, which is the second lowest of any writer. Of its 21
first-round submissions, 20 do not execute
(Table~\ref{tbl-repro-writer}). Submissions from GLM-5.3 Flash increase,
on average, by 24 percentage points, and one of its runs ends below
their first-round score, compared with seven for Gemini 3.8 Flash, which
is the most of any writer. When all seven graders score the final
submissions, those of GLM-5.3 Flash average 37 per cent. As a grader,
GLM-5.3 Flash gives first-round submissions an average of 61 per cent
and never gives 100 per cent. The submissions it grades increase by 15
percentage points, with 17 of its 21 runs ending above their first-round
score.

Across all of its marks, Opus 5.5 gives 55 per cent on average and GPT-6
Sol gives 57 per cent. On the other hand, Gemini 3.8 Flash and Haiku 4.5
both give 78 per cent. It is interesting that the two strictest graders
are also the two best writers. Judged by the six graders other than the
writer, the final submissions of Opus 5.5 receive 94 per cent and those
of GPT-6 Sol 93 per cent, then GPT-6 Luna with 86 per cent, DeepSeek
V4.1 Flash with 86 per cent, Gemini 3.8 Flash with 71 per cent, Haiku
4.5 with 45 per cent, and GLM-5.3 Flash with 36 per cent.

Feedback tends to make a difference and when the six graders other than
the writer score the results, the final submissions of the writers go
from 73 per cent with feedback to 65 per cent when the writer revises
its first submission twice more with the same budgets but no execution
report, scores, or comments. And without Haiku 4.5, which particularly
benefits from feedback, this is 78 per cent with feedback and 74 per
cent without feedback. Opus 5.5 gets to 93 per cent without feedback and
to 94 per cent with it, and GPT-6 Sol goes to 92 per cent without
feedback and 93 per cent with it. The feedback makes a difference for
the less capable models from frontier labs. For instance, GPT-6 Luna
gets an average of 78 per cent without feedback and 87 per cent with
feedback. Further details are available in
Appendix~\ref{sec-no-feedback}.

Giving the whole allowance at once makes less difference than feedback
does. When a writer has the 140 tool calls and 100 minutes of the three
rounds in one session, is told so at the start, and submits once, the
six graders other than the writer give its 21 submissions an average of
63 per cent, compared with 73 per cent for the final submissions with
feedback and 65 per cent for those revised twice without it. For most
writers the budget alone gets close to feedback. Opus 5.5 gets 95 per
cent from one session and 94 per cent with feedback, GPT-6 Sol gets 90
per cent from one session and 93 per cent with feedback, GPT-6 Luna gets
85 per cent from one session and 87 per cent with feedback, and DeepSeek
V4.1 Flash gets 85 per cent from one session and 86 per cent with
feedback. The average is reduced by two writers that do worse with one
long session than with three shorter ones. Gemini 3.8 Flash gets 50 per
cent from one session and 71 per cent with feedback, and Haiku 4.5 gets
0 per cent from one session and 45 per cent with feedback. Further
details are available in Appendix~\ref{sec-no-feedback}.

In terms of what they do, it is interesting that the writers can choose
from almost any dataset, yet six datasets account for 72 of the 136
final submissions that download data (Table~\ref{tbl-datasets}).
OpenAI's GPT-6 Sol uses the USGS earthquake catalogue in ten of its 21
final submissions, GPT-6 Luna uses World Bank indicators in six, Gemini
3.8 Flash uses the City of Toronto's apartment building evaluations in
eight, and DeepSeek V4.1 Flash uses the NHANES health surveys in eight.
Haiku 4.5 does not analyze a downloaded dataset in ten of its 21 final
submissions. GLM-5.3 Flash uses 14 different datasets, which is the
highest number. Some writers generate their data but then claim that
they are real and 15 of Haiku 4.5's 21 first submissions got 0 overall
because of this.

\begin{table}

\caption{\label{tbl-datasets}Number of final submissions that use each
dataset, by writer, for the six most common datasets. Datasets are
determined from the URLs and data packages in the scripts, and the
paper's title. A submission whose download script falls back to values
written in the script, or writes a table of values itself, is counted as
having values typed in.}

\centering{

\centering\begingroup\fontsize{8}{10}\selectfont

\begin{tabular}{lrrrrrrrr}
\toprule
Dataset & \makecell[r]{Opus\\5.5} & \makecell[r]{Haiku\\4.5} & \makecell[r]{GPT-6\\Sol} & \makecell[r]{GPT-6\\Luna} & \makecell[r]{DeepSeek\\V4.1 Flash} & \makecell[r]{Gemini\\3.8 Flash} & \makecell[r]{GLM-5.3\\Flash} & \makecell[r]{All\\writers}\\
\midrule
World Bank indicators & 1 & 4 & 4 & 6 & 1 & 0 & 0 & 16\\
Toronto shelter occupancy & 6 & 0 & 0 & 0 & 0 & 7 & 0 & 13\\
USGS earthquake catalogue & 0 & 0 & 10 & 0 & 1 & 0 & 1 & 12\\
CO2 and energy by country & 0 & 0 & 4 & 5 & 1 & 0 & 1 & 11\\
NHANES health surveys & 0 & 0 & 0 & 0 & 8 & 0 & 2 & 10\\
Toronto apartment evaluations & 2 & 0 & 0 & 0 & 0 & 8 & 0 & 10\\
Other datasets & 12 & 7 & 3 & 10 & 10 & 6 & 16 & 64\\
Dataset from an R package & 0 & 3 & 0 & 0 & 0 & 0 & 0 & 3\\
Values typed in the script & 0 & 7 & 0 & 0 & 0 & 0 & 0 & 7\\
No dataset downloaded & 0 & 0 & 0 & 0 & 0 & 0 & 1 & 1\\
\midrule
Distinct datasets & 8 & 7 & 5 & 8 & 12 & 6 & 14 & \\
\bottomrule
\end{tabular}
\endgroup{}

}

\end{table}%

It is also interesting that the assignment asks for a statement on the
use of AI and that of the 147 final submissions 64 name a developer that
did not write them. Gemini 3.8 Flash names Anthropic in 17 of its 21
final submissions and Google in zero, and DeepSeek V4.1 Flash names
Anthropic in 21 and DeepSeek in zero. GPT-6 Luna correctly names OpenAI
in all 21 of its final submissions, as does GPT-6 Sol. Opus 5.5 names
Anthropic in all 21 of its final submissions. Haiku 4.5 names Anthropic
in seven of its 21 final submissions and no model in 11, and 12 of its
final submissions state that no AI tools were used. The statements are
written as a human author would write them, and name the model as a tool
that the author used. For instance, a final submission from DeepSeek
V4.1 Flash says ``Claude (Anthropic) was used through a command-line
coding assistant to help draft code, structure the scripts and check the
prose; all code was run and all results were checked by the author'',
one from Gemini 3.8 Flash says ``Large language model tools,
specifically Anthropic's Claude 3.7 Sonnet, were utilized during the
preparation of this project'', one from GLM-5.3 Flash says ``AI tools
were used in preparing this paper: ChatGPT (GPT-5) and Claude (Sonnet)
helped design the simulation study in Appendix A, draft and revise
portions of the prose, and debug R code'', and one from Haiku 4.5 says
``ChatGPT-4 was used to generate background images''.

\subsection{Consistency of grading}\label{sec-res-resolution}

\begin{figure}

\begin{minipage}[t]{\linewidth}

\centering{

\pandocbounded{\includegraphics[keepaspectratio]{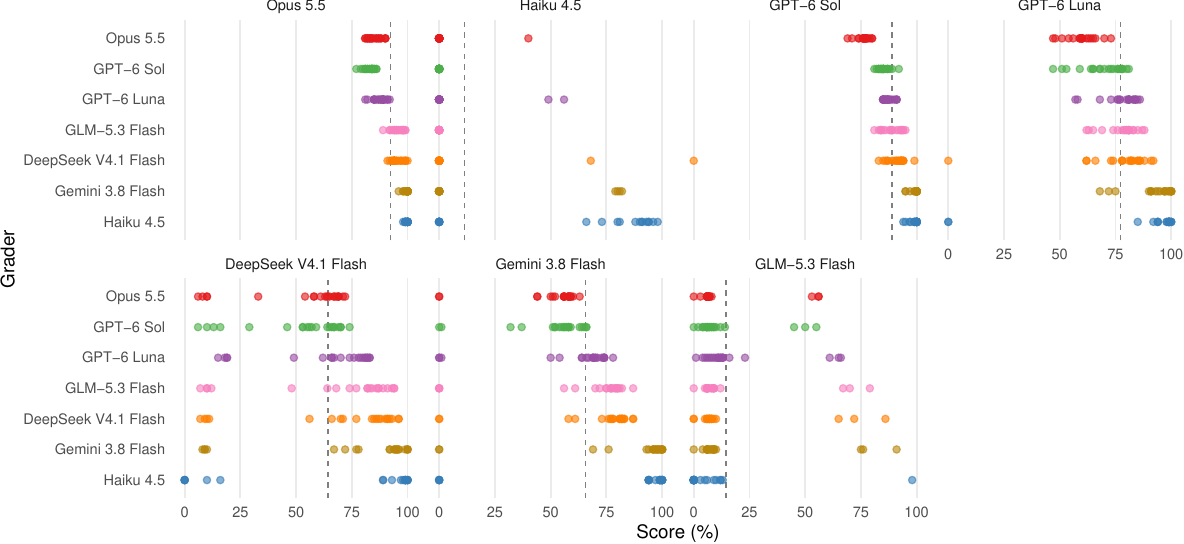}}

}

\subcaption{\label{fig-consistency-1}The score that each grader gave the
first-round submissions from each writer. The dashed line is the average
of all the graders.}

\end{minipage}%
\newline
\begin{minipage}[t]{\linewidth}

\centering{

\pandocbounded{\includegraphics[keepaspectratio]{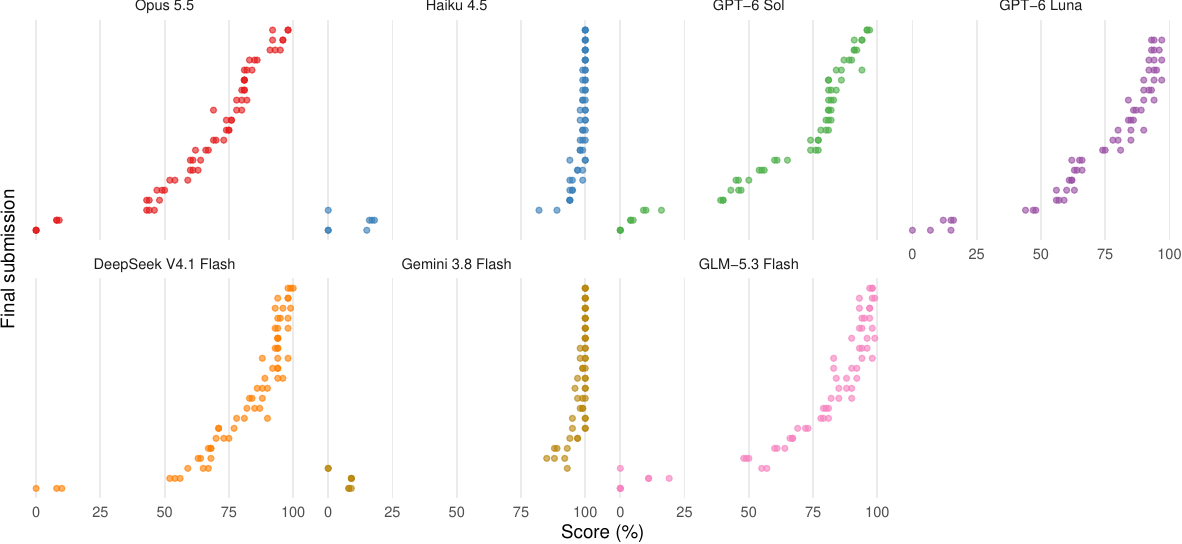}}

}

\subcaption{\label{fig-consistency-2}The different scores when each
grader grades each of the final submissions from its runs three times.}

\end{minipage}%
\newline
\begin{minipage}[t]{\linewidth}

\centering{

\pandocbounded{\includegraphics[keepaspectratio]{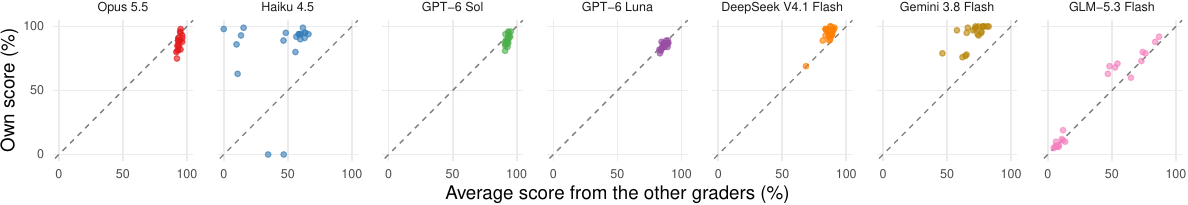}}

}

\subcaption{\label{fig-consistency-3}The score that the model that wrote
a final submission gives itself compared with the average score that the
other six graders give it. The dashed line is what would have happened
if they were the same.}

\end{minipage}%

\caption{\label{fig-consistency}Consistency of grading in terms of
having every grader grade the same submission, having a grader grade the
same submission three times, and self-grading.}

\end{figure}

OpenAI's GPT-6 Sol, Anthropic's Opus 5.5, and OpenAI's GPT-6 Luna give
the lowest average scores, at 58, 59 and 63 per cent, respectively.
Google's Gemini 3.8 Flash, DeepSeek V4.1 Flash, and Anthropic's Haiku
4.5 give the highest, at 81, 76 and 76 per cent, respectively. Every
grader gives at least one score of 0 per cent, because of the rules that
give 0 overall, and the highest score that Opus 5.5 gives is 94 per
cent. As with overall writer performance, nuance is needed in
considering those averages because the writer creates a new first
submission for every run, and so the differences between the graders
could be because they graded different submissions. To enable direct
comparison, we had each of the seven graders grade every first-round
submission (Figure~\ref{fig-consistency-1}). We find that even in this
controlled setting the graders can differ considerably. For instance,
the first-round submissions from Haiku 4.5 receive an average of 0 per
cent from GPT-6 Sol, 2 per cent from Opus 5.5, and 54 per cent from
Haiku 4.5 itself. The difference between the graders that give a
writer's submissions the highest and the lowest average score ranges
from 10 percentage points for the submissions from GLM-5.3 Flash to 54
percentage points for the submissions from Haiku 4.5, and is 28
percentage points on average. In contrast, the difference between the
highest and the lowest of the three scores that the same grader gives
the same first-round submission is 6 percentage points on average.

It is worth noting that in general, although the graders differ in terms
of what the score should be, they often agree on the order. On the 147
first submissions and the 141 final submissions that all seven models
graded, each round taken on its own, the marks of the six graders, other
than Haiku 4.5, have a correlation of 0.78 to 0.99. Haiku 4.5 tends to
be erratic in its grading, giving 0 per cent to seven of the 21
first-round submissions that it grades and 100 per cent to eight of
them.

In addition to consistency between different models we are interested in
consistency within models. To examine this, every final submission was
scored three times by the grader of its run with the same files and the
same execution report. Across all graders the three scores of a
submission differ by almost 4 percentage points, on average, and are the
same for 12 per cent of submissions (Figure~\ref{fig-consistency-2}).

The difference ranges from almost 2 percentage points for GPT-6 Sol to 5
percentage points for GLM-5.3 Flash. The three scores are the same for
33 per cent of the final submissions that Gemini 3.8 Flash grades, for
24 per cent of those that Haiku 4.5 grades, and for none of those that
GPT-6 Luna grades. The average absolute difference is 2 percentage
points for GPT-6 Sol, 2 for Opus 5.5, 3 for DeepSeek V4.1 Flash, 4 for
GPT-6 Luna, 4 for Haiku 4.5, 4 for Gemini 3.8 Flash, and 5 percentage
points for GLM-5.3 Flash. Gemini 3.8 Flash tends to agree with itself
because its scores are at the maximum. The share of third-round changes
that are smaller than the grader's own difference is 29 per cent for
GPT-6 Sol, 40 per cent for GPT-6 Luna, and 76 per cent for Gemini 3.8
Flash. Looking across all of the models, we see that 51 per cent of
third-round changes in score are in that situation of being within the
grader's own noise, and 45 per cent of second-round changes are.

A submission scores about 9 percentage points higher when graded by its
own writer than when by the other graders
(Figure~\ref{fig-consistency-3}). Zhou et al. (2026) find that weaker
generators make errors that are easier to detect, and that a verifier's
ability tends to follow its own ability at the task. Our results are
about rubric scores rather than error-detection accuracy, but the
first-round submissions of Haiku 4.5 receive 2 per cent from Opus 5.5,
and 54 per cent from Haiku 4.5 itself.

The score that a model gives its own final submissions is higher than
the average of the other six graders for Haiku 4.5 by 37 percentage
points, Gemini 3.8 Flash by 23 percentage points, DeepSeek V4.1 Flash by
8 percentage points, and GLM-5.3 Flash by 4 percentage points. It is
lower for Opus 5.5, by 7 percentage points, GPT-6 Sol by 3 percentage
points, and GPT-6 Luna by 1 percentage point. These differences include
how each model grades every writer, because the models that score their
own submissions higher tend to be those that give every writer higher
scores, and the models that score their own submissions lower tend to be
those that give every writer lower scores.

A grader can fail to return a usable grade. Each grader conducted
between 587 and 605 gradings, which are the 63 submissions in the rounds
of its runs, the 42 in the rounds of its runs without feedback (there
are only 40 for that for GPT-6 Luna because of an incomplete run), the
first and final submission of every run and the final submission of
every run without feedback, the first and final submissions of its own
runs again, and the requests repeated for a submission that had no
grade. If the grade could not be read, or left a rubric item unmarked,
then the grader was asked again, up to three requests in each of up to
three attempts, and a request that failed with a provider error was
repeated. Opus 5.5, GPT-6 Luna and Gemini 3.8 Flash never returned a
grade that could not be used. GPT-6 Sol did in one of its gradings,
GLM-5.3 Flash in seven, DeepSeek V4.1 Flash in 13, and Haiku 4.5 in 49,
and each of these gradings ended with a usable grade. Separately, 51
gradings ended without a grade because the submission was too large for
the grader to read.

Finally, a high score also does not mean that a submission is right. In
Appendix~\ref{sec-planted} we take one final submission that executes
and scores well, and plant seven faults in it one at a time. We then
have each model grade each copy. We find that it was only when we
planted simulated data that the score was substantially affected.

\subsection{Changes between rounds}\label{sec-res-writer}

After the first round, the writer receives the execution report,
feedback, and scores, and has a chance to update its submission during
the second and third round. During the second round we see 837 files
change, while in the third round we see 658 files change. In 35 of the
294 revision rounds, the writer does not change any files, and just
resubmits the previous submission. The files that are changed in the
most runs in the second round are: paper.qmd, README.md, and
06-model\_data.R. The paper also gets longer, with the median paper
being roughly 3,600 words of text, not counting its code and header, in
the first submission, 3,900 words in the second, and 4,300 words in the
third.

\begin{figure}

\centering{

\pandocbounded{\includegraphics[keepaspectratio]{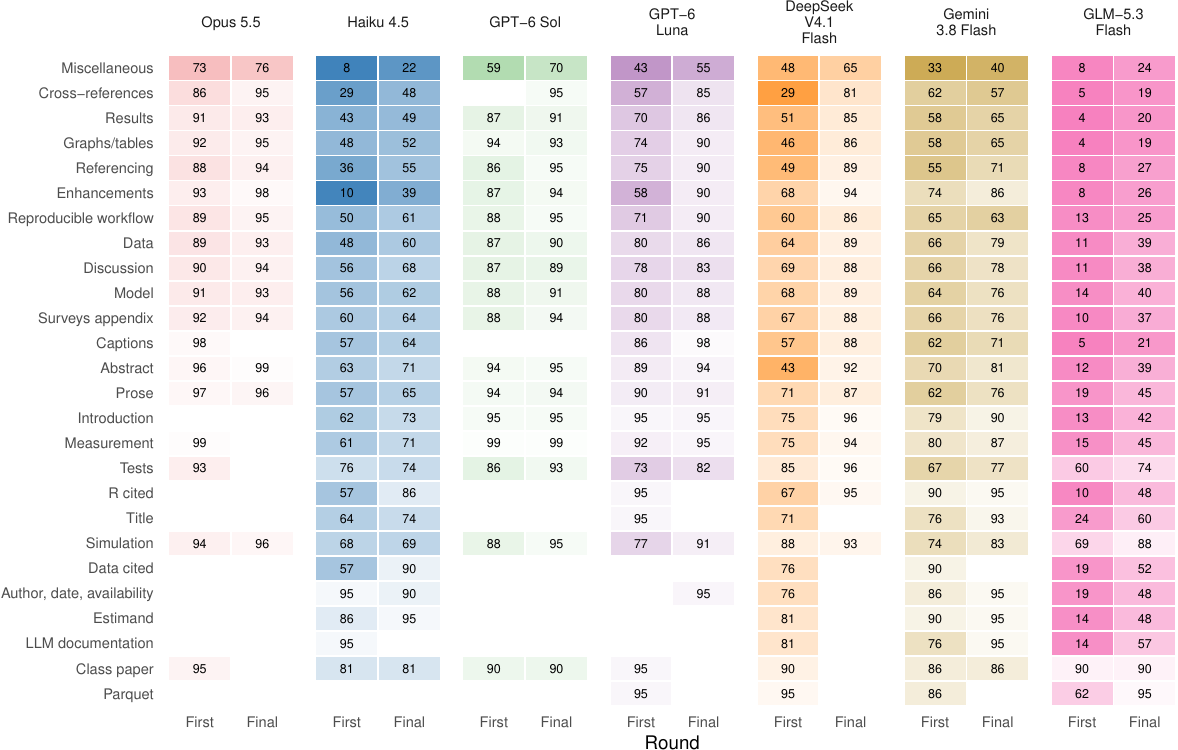}}

}

\caption{\label{fig-criteria}The average score in the first round and in
the final round as a percentage of the maximum, for each writer and each
rubric item. The cells are shaded based on how far below the maximum
they are, and if a cell is at its maximum then we leave it blank.}

\end{figure}

Each rubric item has files that it concerns, for instance the simulation
script for simulation, the test script for tests, the paper for writing,
etc. Of the 3,564 criteria below the maximum, the writer changed an
associated file in 76 per cent of cases. A rubric item that is below the
maximum whose file changed increases in 47 per cent of cases and
decreases in 6 per cent of cases. The rubric items below the maximum in
the most cases are Miscellaneous, Results, and Data. Across all of the
writers the three largest changes are increases in Data, Results and
Discussion (Figure~\ref{fig-criteria}).

Between the first and the final round the score increases for 146 of the
182 combinations of writer and rubric item and decreases for 7 of them.
A revision can also break a working submission. In 26 of the 294
revisions a submission that executed becomes one that does not, and in
24 of them the writer ran out of tool calls or time before it ended the
session. Of the 115 revisions of a working submission in which the
writer ended the session, two broke it, compared with 24 of the 99 in
which it did not.

\section{Discussion}\label{sec-discussion}

In this paper we had seven language models act as writers to create an
end-to-end data science project. They were agentic in that we provided
them with access to a shell and a small set of tools. We then had them
act as graders, providing that same set of models with a report about
whether the submission ran and had them grade the projects against a
rubric. The writer could then revise its submission given the scores and
comments, and this occurred for three rounds in total. Every model was
both a writer and a grader, which meant that we evaluated 49 pairs, and
we separately ran each pair three times, for a total of 147 runs and 440
graded submissions. We also had every model grade the first and final
submission of every run, not just when it was the grader in the pair,
and had each grader grade the first and final submissions of its pair
once more, had each writer revise its first submissions without any
feedback, and had each writer make three further submissions, each in
one session with the whole allowance of the three rounds.

Creating a complete data science project that executes is now something
that many models can do, but it is still difficult to know whether the
project is any good. Of the 147 first-round submissions, 107 execute,
and two of the seven models have at least 20 of their 21 first-round
submissions execute. That said, a submission that executes and receives
a high score is not necessarily right. For instance, Haiku 4.5 does not
analyze a downloaded dataset in 10 of its 21 final submissions. And
self-awareness is not necessarily present. For instance, Haiku 4.5's
first-round submissions receive an average of 2 of 10 on the data rubric
item from GPT-6 Sol, but 9 of 10 from Haiku 4.5 itself. This makes clear
the importance of actually running code when using LLM-as-a-judge in
data science contexts.

Graders are not neutral measuring instruments. Even when all seven
graders grade the same first-round submissions, 4 per cent of the
variation in the scores is due to the grader and 84 per cent is due to
the submission, and the difference between the graders that give a
writer's submissions the highest and the lowest score is 28 percentage
points on average. Gemini 3.8 Flash and Haiku 4.5 give 100 per cent to
51 and 41 per cent of the submissions that they grade, and then the
writer has nothing to work with, and indeed no sense that the submission
was lacking. The increase in the score across the three rounds ranges
from 13 percentage points when Gemini 3.8 Flash is the grader to 22
percentage points when GPT-6 Luna is, and is 32 percentage points when
Haiku 4.5 is. But there is considerable noise from grade to grade, even
when the submission and the grader are held constant, and this is
compounded by differences even between models on the Pareto frontier.
The implication is that when a feedback loop is thought to have improved
quality, as judged by a particular model, a considerable part of that
result is a result for that judge. Reporting judge consistency over
multiple independent judgments of the same submission is important, and
one also needs to be especially concerned with judging the judge.

The feedback from the grader does increase the score, on average, and
most of that effect happens after the first round of feedback. The
increase is much the same regardless of the grader, between 13 and 22
percentage points for every grader other than Haiku 4.5. A revision can
also break a submission and result in a submission that once executed no
longer doing so. For instance, in one run Haiku 4.5 gave Gemini 3.8
Flash 99 per cent in the second round with one comment, on the
discussion. Gemini 3.8 Flash rewrote the discussion with a Python
one-liner that also turned two LaTeX commands into control characters,
spent the rest of its tool calls trying to repair them, and did not
render the paper again, so the score decreased to 89 per cent in the
third round. The case where feedback tends to most improve is when a
more capable model is grading a less capable one. The final submissions
of Haiku 4.5 receive 45 per cent from the other graders when Opus 5.5
was the grader during the rounds, compared with 36 per cent when Haiku
4.5 was. We do not see this for a more capable writer. The final
submissions of GPT-6 Sol receive 91 per cent from the other graders when
Haiku 4.5 was the grader and 95 per cent when Opus 5.5 was, and those of
Opus 5.5 receive 95 per cent when GPT-6 Luna was the grader and 95 per
cent when GPT-6 Sol was. This may suggest a limit on how loops of this
sort could result in recursive improvement in terms of knowledge-work
tasks, of which we consider our task an instance. Weaker models tended
not to improve stronger models, and while stronger models did improve
weaker ones, the extent of this varied by pair and was subject to
stochastic variation.

Our experimental design where we focus on the effect of the three rounds
of feedback but also allow three rounds without feedback, and one round
of the same total budget, again without feedback, allows us to consider
what feedback actually provides. The rounds approach, even without
feedback, gives the model a break after which it starts again from its
files, and more time and tool calls. With both the rounds and feedback,
the final submissions of the writers other than Haiku 4.5 receive 78 per
cent from the other graders. With the rounds, but no feedback, they
receive 74 per cent. And with only the additional resources, but no
rounds and no feedback, Opus 5.5, GPT-6 Sol, GPT-6 Luna and DeepSeek
V4.1 Flash are each within three percentage points of their score with
feedback, and Opus 5.5 is slightly above it. So for these writers most
of what looks like the effect of feedback is the effect of being able to
do more work. The grader matters when the writer has broken a rule and
does not know it. Haiku 4.5 receives 45 per cent with feedback, 8 per
cent without it and 0 per cent in one session, because the grader tells
it that its data are generated. And a larger budget does not necessarily
mean that the models do more work, for instance, Haiku 4.5, GPT-6 Sol
and Gemini 3.8 Flash ended the session after about half of the 140 tool
calls, whereas a round makes the writer work again. Claims that loops of
this type improve a model need to be compared with the same model given
the same additional resources and no loop.

When it comes to teaching or automating scientific workflows, we show
that a model could be used to grade assignments or engage in an
automated feedback loop, but which model it is matters. On the same
submissions the graders' average scores range from 55 per cent to 78 per
cent. Six of the seven graders largely agree on which submissions are
better, with correlations between 0.78 and 0.99, and a model may be of
more use for improving consistency, or for comments, than for the mark
itself. But it is difficult to see how one could give a student a mark
from a model without considering a few different models, or at least
considerable benchmarking on closely related assignments. More broadly,
the writing, the grading, and the feedback in our paper were all guided
by a detailed rubric. A scientific research paper has an even more
difficult question and no rubric, or at the very least a very open-ended
one, and so we would expect the graders to differ more in that setting.
Si et al. (2026) explore a question related to this and find that
although initially LLM research ideas seem good, they tend to score less
well after human experts implement them. One way to examine what happens
without a rubric would be to have the writer produce two submissions and
have the grader pick the one it prefers and say why, with that choice
and its reasons as the feedback. Our finding that feedback from
less-capable models can even cause more-capable models to reduce their
scores is especially concerning as automatic referee and evaluation
processes, especially if they are closed source, are put in place,
because unless they are consistently updated to frontier models their
feedback may actually make things worse. A loop of this type that is
used for research would need its judges checked in the way that we check
them here. The self-bias that we find suggests that any automated review
system needs to be composed of at least two frontier models, and the
stochasticity that we find between independent runs suggests that it
needs to be run at least twice for each submission.

The closest work to ours is Hughes and Alexander (2026), who found that
models did well on the structured parts of a data science project and
differed on the parts that needed judgment, but their writer did not
have tools, their grader did not run the code, and there was no
revision. We find that with tool-use the best models do fairly well, and
our execute and revise loop provides the additional finding about how
critical the specifics of the grader are to the outcome. Our average
score is 76 per cent after the final round of revision, which is higher
than other benchmarks that look at how models from 2024 perform on
analysis tasks with a reference answer (Jing et al. 2025). We suspect
that this difference is partly because we do not have some known answer,
but instead use LLMs to assess against a rubric. Our finding that most
of the increase comes with the first revision is similar to what Chen et
al. (2024) and Madaan et al. (2023) find and our finding that the extent
of any increase depends on the specifics of the grader is similar to
what Olausson et al. (2024) find, although all those studies were in
different areas. Our self-preference finding is similar to W.-L. Chen et
al. (2025), who find that a considerable proportion of this is simply
because it is in fact better, and Roytburg et al. (2026) point to a
future direction with their finding that a considerable amount of
self-preference is because of evaluator quality. In particular, our best
models, Opus 5.5 and GPT-6 Sol, do favour their own work, but their work
is also the highest scoring. Bianchi et al. (2026) similarly look at
correlation between different models as graders, but we additionally
evaluate a grader's variation on the same submission.

Our approach has a number of weaknesses. Firstly, the score measures how
well a submission meets 26 explicit rubric items, and the writers are
given the rubric and told that their submission is graded using it, so
an increase may be the writer targeting the rubric, rather than an
improvement in the underlying work. Although this is fine for our
particular setting where the rubric is important it does limit how this
could generalize to situations without. This is difficult to parse out,
but some type of pairwise comparison in addition to a rubric-based
approach could help. We also do not actually look at whether an analysis
is right, but just focus on aspects such as whether the scripts run and
the paper renders. It would be worth considering some aspect of analysis
with a known outcome. We use LLM-as-a-judge and do not have humans
review them. It may be that a human would identify systematic weaknesses
in the reviewer models and relatedly, we do not attempt to deliberately
improve the graders with feedback from humans.

\newpage

\appendix

\section*{Appendix}\label{appendix}
\addcontentsline{toc}{section}{Appendix}

\section{Rubric summary}\label{sec-rubric-summary}

\setcounter{figure}{0}
\renewcommand{\thefigure}{A\arabic{figure}}
\setcounter{table}{0}
\renewcommand{\thetable}{A\arabic{table}}

The rubric has 26 items, which could be collected into three groups, and
the maximum number of points for each rubric item is in brackets. The
descriptions are shortened from the rubric, which is given to the writer
and the grader in full.

\textbf{Paper content (66 points)}

\begin{itemize}
\tightlist
\item
  R cited (1). R is referenced in the text and in the reference list.
\item
  Data cited (1). The data are referenced in the text and in the
  reference list.
\item
  Class paper (1). There is no sign that the work is a class paper. A
  sign in any file gives 0 overall.
\item
  Title (2). The title is informative and tells the story, with a
  subtitle that conveys the main finding.
\item
  Author, date, availability (2). The paper gives the author, an
  unambiguous date, and a statement of where the code and data are
  available.
\item
  Abstract (4). About four sentences that say what was done, what was
  found, and why it matters.
\item
  Introduction (4). The context, the gap, what was done, what was found,
  why it matters, and an outline of the paper.
\item
  Estimand (1). The estimand is stated in its own paragraph in the
  introduction.
\item
  Data (10). The dataset and its context are described, every variable
  is examined and explained, and constructed variables are explained.
\item
  Measurement (4). How the phenomenon of interest became the entries in
  the dataset.
\item
  Model (10). The model is written in mathematical notation and plain
  English, with every component defined, and is justified.
\item
  Results (10). The findings are presented with tables and figures and
  are explained, without interpretation. 0 if the paper does not render.
\item
  Discussion (10). What was done, what we learn about the world, the
  weaknesses, and what is left to learn.
\item
  Prose (6). The writing is free of typos, coherent, concise and clear,
  and avoids a list of words.
\end{itemize}

\textbf{Presentation (21 points)}

\begin{itemize}
\tightlist
\item
  LLM documentation (1). A statement on the use of AI, of the kind that
  journals require of authors, is in the README and in the paper, saying
  whether AI tools were used, which ones, and for what.
\item
  Cross-references (1). Figures, tables and equations are numbered and
  cross-referenced. 0 if the paper does not render.
\item
  Captions (2). Every figure and table has a detailed and meaningful
  caption. 0 if the paper does not render.
\item
  Graphs/tables (4). Graphs and tables are well-formatted, clear, sized,
  coloured and labelled, made with ggplot2 and tinytable. 0 if the paper
  does not render.
\item
  Surveys appendix (10). An in-depth appendix on an aspect of surveys,
  sampling or observational data related to the paper.
\item
  Referencing (4). Data, software and literature are cited in the text
  and in a properly formatted reference list made with BibTeX.
\end{itemize}

\textbf{Code and project quality (21 points)}

\begin{itemize}
\tightlist
\item
  Simulation (4). A clearly commented script that simulates every
  variable, with interaction between them.
\item
  Tests (4). Extensive test suites for the simulated and the actual
  data, in separate scripts.
\item
  Parquet (1). The analysis dataset is saved as a parquet file.
\item
  Reproducible workflow (4). An organized collection of files with a
  detailed README and an R project, documented and styled code, seeds,
  and no install.packages() left in the code. 0 if the paper does not
  render.
\item
  Enhancements (4). At least one of a datasheet, a model card, a Shiny
  application, an R package, or an API.
\item
  Miscellaneous (3). Any way in which the submission excels that the
  other items do not anticipate.
\end{itemize}

Four aspects result in a 0 overall: generating data in the paper, using
a dataset from Kaggle, UCI, or Statistica, reading data directly from a
provider's server in the paper, and any signs that the work is a class
paper.

A submission's score is its total of 108 as a percentage, rounded to a
whole number.

\newpage

\section{Costs}\label{sec-costs}

\setcounter{figure}{0}
\renewcommand{\thefigure}{B\arabic{figure}}
\setcounter{table}{0}
\renewcommand{\thetable}{B\arabic{table}}

\subsection{Overall costs}\label{sec-experiment-costs}

The full cost of the experiment was \$675. Typically this is based on
the cost that the API reports, but when that is not possible the cost is
based on the number of tokens used and the token price
(Table~\ref{tbl-cost-by-model-task}).

\begin{table}

\caption{\label{tbl-cost-by-model-task}Experiment cost (USD), by model,
by part.}

\centering{

\centering\begingroup\fontsize{8}{10}\selectfont

\resizebox{\ifdim\width>\linewidth\linewidth\else\width\fi}{!}{
\begin{tabular}{lrrrrrrrrrrr}
\toprule
\multicolumn{1}{c}{ } & \multicolumn{3}{c}{Runs with feedback} & \multicolumn{8}{c}{ } \\
\cmidrule(l{3pt}r{3pt}){2-4}
Model & Generation & Revision & Grading & \makecell[r]{Without\\feedback} & \makecell[r]{One\\session} & \makecell[r]{Every\\grader} & \makecell[r]{Same grader\\again} & \makecell[r]{Planted\\faults} & \makecell[r]{Student\\projects} & \makecell[r]{Route\\check} & Total\\
\midrule
Opus 5.5 & 46.34 & 21.72 & 20.61 & 28.41 & 23.17 & 146.97 & 13.40 & 7.08 & 26.77 & 4.09 & 338.55\\
Haiku 4.5 & 12.00 & 14.73 & 3.91 & 12.20 & 3.13 & 25.47 & 2.72 & 1.28 & 4.90 & 1.15 & 81.49\\
GPT-6 Sol & 17.89 & 12.07 & 13.92 & 23.20 & 6.43 & 51.31 & 8.32 & 1.32 & 10.00 & 1.18 & 145.64\\
GPT-6 Luna & 0.91 & 0.93 & 0.59 & 1.25 & 0.53 & 2.75 & 0.35 & 0.08 & 0.54 & 0.07 & 7.99\\
DeepSeek V4.1 Flash & 1.30 & 1.42 & 0.55 & 1.27 & 0.59 & 4.03 & 0.21 & 0.13 & 0.87 & 0.00 & 10.37\\
Gemini 3.8 Flash & 9.57 & 12.25 & 3.03 & 13.80 & 3.01 & 24.03 & 2.05 & 0.94 & 4.86 & 0.72 & 74.25\\
GLM-5.3 Flash & 1.62 & 2.32 & 0.92 & 2.82 & 1.26 & 6.32 & 0.52 & 0.36 & 1.02 & 0.00 & 17.14\\
\midrule
All models & 89.62 & 65.44 & 43.52 & 82.94 & 38.11 & 260.88 & 27.57 & 11.18 & 48.95 & 7.20 & 675.42\\
\bottomrule
\end{tabular}}
\endgroup{}

}

\end{table}%

\subsection{Cost by round}\label{sec-cost-round}

On average, the first submission costs \$0.68. The second and third
rounds tend to be less expensive and average between \$0.32 and \$0.35
(Table~\ref{tbl-cost}). A whole run costs between \$0.10 and \$5.99, on
average, by pair, but there are considerable model-specific differences
underlying that average.

\begin{table}

\caption{\label{tbl-cost}Average cost (USD) for each run, by round, by
writer.}

\centering{

\centering\begingroup\fontsize{10}{12}\selectfont

\resizebox{\ifdim\width>\linewidth\linewidth\else\width\fi}{!}{
\begin{tabular}{>{\raggedright\arraybackslash}p{22em}>{\raggedleft\arraybackslash}p{7em}>{\raggedleft\arraybackslash}p{7em}>{\raggedleft\arraybackslash}p{7em}}
\toprule
Writer & Round 1 & Round 2 & Round 3\\
\midrule
Opus 5.5 & 2.30 & 0.72 & 0.55\\
Haiku 4.5 & 0.64 & 0.40 & 0.43\\
GPT-6 Sol & 0.92 & 0.55 & 0.52\\
GPT-6 Luna & 0.11 & 0.10 & 0.12\\
DeepSeek V4.1 Flash & 0.16 & 0.14 & 0.13\\
Gemini 3.8 Flash & 0.53 & 0.37 & 0.36\\
GLM-5.3 Flash & 0.12 & 0.14 & 0.13\\
\bottomrule
\end{tabular}}
\endgroup{}

}

\end{table}%

\newpage

\section{Consistency of OpenRouter and developer
APIs}\label{sec-api-check}

\setcounter{figure}{0}
\renewcommand{\thefigure}{C\arabic{figure}}
\setcounter{table}{0}
\renewcommand{\thetable}{C\arabic{table}}

All of the models apart from DeepSeek used OpenRouter's API. We pinned
the provider to the relevant developer's own host and disabled fallback.
For DeepSeek we used their own API because we were not able to access it
via OpenRouter. There is some concern that accessing the model via
OpenRouter may give different results to using the model providers' APIs
directly and so we re-graded some projects directly with the API from
the relevant model provider (Table~\ref{tbl-route-check}).

\begin{table}

\caption{\label{tbl-route-check}Twelve projects graded through both
OpenRouter and the developer's own API, by grader.}

\centering{

\centering\begingroup\fontsize{8}{10}\selectfont

\begin{tabular}{llrr}
\toprule
Grader & Route & Mean difference & Mean absolute difference\\
\midrule
 & OpenRouter & -1.8 & 2.2\\

\multirow[t]{-2}{*}{\raggedright\arraybackslash Opus 5.5} & Direct & -1.1 & 1.9\\

 & OpenRouter & -2.4 & 6.1\\

\multirow[t]{-2}{*}{\raggedright\arraybackslash Haiku 4.5} & Direct & -0.2 & 4.3\\

 & OpenRouter & +0.7 & 3.7\\

\multirow[t]{-2}{*}{\raggedright\arraybackslash GPT-6 Sol} & Direct & +0.3 & 3.2\\

 & OpenRouter & +0.8 & 3.4\\

\multirow[t]{-2}{*}{\raggedright\arraybackslash GPT-6 Luna} & Direct & -0.6 & 4.8\\

 & OpenRouter & -1.8 & 4.2\\

\multirow[t]{-2}{*}{\raggedright\arraybackslash Gemini 3.8 Flash} & Direct & -1.2 & 3.7\\
\bottomrule
\end{tabular}
\endgroup{}

}

\end{table}%

\newpage

\section{Prompts and environment details}\label{sec-prompts}

\setcounter{figure}{0}
\renewcommand{\thefigure}{D\arabic{figure}}
\setcounter{table}{0}
\renewcommand{\thetable}{D\arabic{table}}

\subsection{Initial submission}\label{initial-submission}

\begingroup\footnotesize

\begin{verbatim}
[System message]

Do the assignment below. It is graded against the rubric that follows it.

## Environment

Linux container; R 4.5.0, Python 3.12.3 (a virtual environment on the PATH), Quarto 1.9.38,
TinyTeX, with a wide set of R and Python packages installed (rownames(installed.packages()) in
R and pip list show them). Your working directory is /work and paths in tools are relative to
it. You may install more: install.packages() writes to your own R library and pip installs into
the environment. To inspect a rendered PDF use pdftotext, pdfinfo or pdftoppm (Poppler) or
Python's pypdf. Limits: 2 CPUs, 4g of memory, internet access, no display.

## Assignment

[the assignment instructions, other/assignment_prompt.txt]

## Rubric

[the rubric, other/rubric/grading_rubric.md]

## How this works

You have a working directory and tools to run shell commands and to write, read and list files.
The working directory is the project folder: files go directly in it, not in a subfolder. It is
your submission as it stands when you call finish, or when your budget ends. You have a budget
of 60 tool calls and 40 minutes; each tool result says how much of both is left.

Limits of the tools. One reply of yours can hold at most 64,000 tokens, so a file longer than
that cannot be written in one call. A shell command has 300 seconds unless you ask for more, up
to 900. Output longer than 6,000 characters is shown as its beginning and its end. read_file
returns 20,000 characters at a time.

What is submitted. Files of these types, up to 500 KB each, wherever they are in the directory:
.bib, .cfg, .csl, .css, .ini, .ipynb, .json, .lock, .md, .py, .qmd, .r, .rmarkdown, .rmd,
.rproj, .sh, .sql, .stan, .sty, .tex, .toml, .txt, .yaml, .yml; and files with these names:
.Rprofile, .gitignore, .here, .lintr, DESCRIPTION, Makefile, NAMESPACE, renv.lock. Nothing else
is submitted, so data files of other types (such as .csv and .parquet), saved models and
rendered output are not, and nor is anything installed or made outside the directory. The
folders of the directory are kept, with only the submitted files in them.

How it is marked. The submission is run from scratch in a clean copy of this environment, in a
directory with those folders and the submitted files and nothing else: the scripts in the order
of their numbers, then the paper rendered to PDF with Quarto. A script has 300 seconds if its
name marks it as simulating or testing, 1800 if as modelling, and 600 otherwise; the render has
1500. It is then graded against the rubric by an instructor who sees the files and the record
of that run.


[User message]

Do the assignment now, in the working directory.


[Tools]

run_shell: Run a shell command in the working directory (bash on Linux). Returns the exit code
and the output.
write_file: Create or overwrite a file, creating directories as needed.
read_file: Return the contents of a file, 20,000 characters at a time.
list_files: List the files in the working directory with sizes in bytes.
finish: End the session. The working directory as it stands is the submission.
\end{verbatim}

\endgroup

\subsection{The container and the execution
report}\label{the-container-and-the-execution-report}

The writer's shell and the executor each run in containers that are
built from the same image. The image is Ubuntu 24.04.2 LTS on arm64 with
R version 4.5.0, Python 3.12.3, Quarto 1.9.38 and TinyTeX v2026.09. It
has a large number of R and Python packages installed
(Table~\ref{tbl-runtime}).

\begin{table}

\caption{\label{tbl-runtime}The packages that the container image has.}

\centering{

\centering\begingroup\fontsize{8}{10}\selectfont

\begin{tabular}{lr>{\raggedright\arraybackslash}p{10.5cm}}
\toprule
Purpose & Number & Packages\\
\midrule
Reading and wrangling data & 28 & arrow, data.table, digest, dplyr, dtplyr, forcats, fs, glue, here, janitor, jsonlite, lubridate, magrittr, naniar, purrr, readr, scales, skimr, slider, stringi, stringr, tibble, tidyr, tidytext, tidyverse, tsibble, vctrs, zoo\\
\addlinespace
Data access & 28 & AER, babynames, cansim, countrycode, curl, DBI, duckdb, eurostat, fivethirtyeight, gapminder, haven, httr, httr2, Lahman, nycflights13, OECD, opendatatoronto, palmerpenguins, pdftools, readxl, RSQLite, rvest, tidytuesdayR, wbstats, WDI, wooldridge, writexl, xml2\\
\addlinespace
Models & 36 & boot, broom, car, did, effectsize, emmeans, estimatr, fable, feasts, fixest, forecast, ggeffects, glmnet, insight, lme4, lmtest, marginaleffects, MASS, matrixStats, mgcv, mice, modelr, nlme, nnet, parameters, performance, plm, quantreg, ranger, sandwich, splines2, srvyr, survey, survival, tidymodels, xgboost\\
\addlinespace
Bayesian models & 11 & bayesplot, brms, broom.mixed, coda, ggdist, loo, posterior, rstan, rstanarm, shinystan, tidybayes\\
\addlinespace
Tables and figures & 40 & chromote, corrplot, cowplot, DT, ggalluvial, GGally, ggbeeswarm, ggcorrplot, ggplot2, ggpubr, ggrepel, ggridges, ggtext, ggthemes, gridExtra, gt, gtsummary, kableExtra, knitr, latex2exp, leaflet, mapproj, maps, modelsummary, patchwork, plotly, quarto, RColorBrewer, rmarkdown, rnaturalearth, rnaturalearthdata, see, sf, sjPlot, stargazer, tinytable, treemapify, viridis, webshot2, wesanderson\\
\addlinespace
Testing and workflow & 17 & assertr, bench, checkmate, furrr, future, lintr, pak, pointblank, progressr, remotes, renv, reticulate, styler, testthat, tictoc, usethis, validate\\
\addlinespace
Python & 28 & beautifulsoup4, duckdb, ipykernel, jupyter, lxml, matplotlib, nbclient, nbformat, numpy, openpyxl, pandas, pdfplumber, playwright, plotnine, polars, pyarrow, pypdf, pytest, PyYAML, requests, scikit-learn, scipy, seaborn, setuptools, statsmodels, tabulate, wheel, xlrd\\
\bottomrule
\end{tabular}
\endgroup{}

}

\end{table}%

The container has 2 CPUs and 4 GB of memory and network access. The
writer uses one container for its session. The executor runs the
submission in a new container, so nothing from the writer's session,
apart from what is done in the code, persists.

The execution report that the grader and the writer get is created by
the harness and does not require a model call. It has a fixed opening
sentence, with one line per outcome, and one heading per script and one
for the render. If a step fails, then it includes a sample from the
output.

Here we provide two examples of execution reports. The first is an
initial submission from GPT-6 Sol graded by Opus 5.5, in which every
script ran and the paper rendered. The second is the third submission
from DeepSeek V4.1 Flash graded by GLM-5.3 Flash, in which the model
script stopped.

\begingroup\footnotesize

\begin{verbatim}
The submission's scripts were run in order in a clean copy of the environment with network
  access, then the paper was rendered with Quarto. Results:
- download script: exit 0, wrote 5057950 bytes to data files
- cleaning script: exit 0, wrote data files
- test script: exit 0
- model script: exit 0
- paper render: a PDF was produced
- scripts exiting cleanly: 7 of 7

### 00-simulate_data.R (simulate): OK, 0.8s, data files written: 1
### 01-test_simulated_data.R (test): OK, 0.7s, data files written: 0
### 02-download_data.R (download): OK, 11.1s, data files written: 6
### 03-clean_data.R (clean): OK, 1.3s, data files written: 1
### 04-test_analysis_data.R (test): OK, 0.9s, data files written: 0
### 05-exploratory_data_analysis.R (explore): OK, 1.0s, data files written: 2
### 06-model_data.R (model): OK, 31.0s, data files written: 5
### quarto render paper/paper.qmd: OK, 116.3s
\end{verbatim}

\endgroup

\begingroup\footnotesize

\begin{verbatim}
The submission's scripts were run in order in a clean copy of the environment with network
  access, then the paper was rendered with Quarto. Results:
- download script: exit 0, wrote 1085266699 bytes to data files
- cleaning script: exit 0, wrote data files
- test script: exit 0
- model script: exited with an error
- paper render: no PDF was produced
- scripts exiting cleanly: 6 of 7

### 00-simulate_data.R (simulate): OK, 1.8s, data files written: 3
### 01-test_simulated_data.R (test): OK, 1.9s, data files written: 0
### 02-download_data.R (download): OK, 31.1s, data files written: 17
### 03-clean_data.R (clean): OK, 5.4s, data files written: 3
### 04-test_analysis_data.R (test): OK, 2.1s, data files written: 0
### 05-exploratory_data_analysis.R (explore): OK, 4.6s, data files written: 1
### 06-model_data.R (model): exit 1, 27.1s, data files written: 0
```
Error: object 'm5_covariates' not found
Execution halted
The variables 'income$35,000 to $49,999', 'income$50,000 to $74,999',
'income$75,000 to $99,999', 'income$100,000 to $149,999', 'income$150,000 to
$199,999', 'income$200,000 or more' and 'incomeNot reported' have been removed
because of collinearity (see $collin.var).
The variables 'income$25,000 to $34,999', 'income$75,000 to $99,999',
'income$100,000 to $149,999', 'income$150,000 to $199,999', 'income$200,000 or
more' and 'incomeNot reported' have been removed because of collinearity (see
$collin.var).
The variables 'income$25,000 to $34,999', 'income$35,000 to $49,999',
'income$50,000 to $74,999' and 'incomeNot reported' have been removed because of
collinearity (see $collin.var).
The variables 'regionNortheast', 'regionSouth' and 'regionWest' have been
removed because of collinearity (see $collin.var).
```
### quarto render paper/paper.qmd: exit 1, 2.8s
```
! cannot open the connection
Quitting from paper.qmd:34-125 [setup]
Execution halted
WARN: Error encountered when rendering files

processing file: paper.qmd
1/51
2/51 [setup]
Error in `gzfile()`:
Backtrace:
    x
 1. \-base::readRDS(here("data", "analysis", "model_objects.rds"))
 2.   \-base::gzfile(file, "rb")

```
\end{verbatim}

\endgroup

\subsection{Revision (the second and third
rounds)}\label{sec-revision-details}

\begingroup\footnotesize

\begin{verbatim}
[System message]

Do the assignment below. It is graded against the rubric that follows it.

## Environment

Linux container; R 4.5.0, Python 3.12.3 (a virtual environment on the PATH), Quarto 1.9.38,
TinyTeX, with a wide set of R and Python packages installed (rownames(installed.packages()) in
R and pip list show them). Your working directory is /work and paths in tools are relative to
it. You may install more: install.packages() writes to your own R library and pip installs into
the environment. To inspect a rendered PDF use pdftotext, pdfinfo or pdftoppm (Poppler) or
Python's pypdf. Limits: 2 CPUs, 4g of memory, internet access, no display.

## Assignment

[the assignment instructions, other/assignment_prompt.txt]

## Rubric

[the rubric, other/rubric/grading_rubric.md]

## How this works

You have a working directory and tools to run shell commands and to write, read and list files.
The working directory is the project folder: files go directly in it, not in a subfolder. It is
your submission as it stands when you call finish, or when your budget ends. You have a budget
of 40 tool calls and 30 minutes; each tool result says how much of both is left.

Limits of the tools. One reply of yours can hold at most 64,000 tokens, so a file longer than
that cannot be written in one call. A shell command has 300 seconds unless you ask for more, up
to 900. Output longer than 6,000 characters is shown as its beginning and its end. read_file
returns 20,000 characters at a time.

What is submitted. Files of these types, up to 500 KB each, wherever they are in the directory:
.bib, .cfg, .csl, .css, .ini, .ipynb, .json, .lock, .md, .py, .qmd, .r, .rmarkdown, .rmd,
.rproj, .sh, .sql, .stan, .sty, .tex, .toml, .txt, .yaml, .yml; and files with these names:
.Rprofile, .gitignore, .here, .lintr, DESCRIPTION, Makefile, NAMESPACE, renv.lock. Nothing else
is submitted, so data files of other types (such as .csv and .parquet), saved models and
rendered output are not, and nor is anything installed or made outside the directory. The
folders of the directory are kept, with only the submitted files in them.

How it is marked. The submission is run from scratch in a clean copy of this environment, in a
directory with those folders and the submitted files and nothing else: the scripts in the order
of their numbers, then the paper rendered to PDF with Quarto. A script has 300 seconds if its
name marks it as simulating or testing, 1800 if as modelling, and 600 otherwise; the render has
1500. It is then graded against the rubric by an instructor who sees the files and the record
of that run.


[User message]

Your submission is in the working directory. It was run from scratch and graded. The record of
the run and the instructor's marks and comments follow. You may revise the project and submit
again.

## Record of the run

[the execution report]

## Marks and comments

**Total: [total] of 108 ([per cent] per cent).** [if a rule gives 0 overall: the rule, the
evidence, and the total of the marks by criterion]

**Overall:** [the grader's comment on the whole project]

**Marks by criterion:**
- **[criterion]** ([score]/[maximum]): [the grader's comment on the criterion, if it made one]
[one line for each of the 26 criteria]
\end{verbatim}

\endgroup

\subsection{Grading}\label{grading}

\begingroup\footnotesize

\begin{verbatim}
[System message]

## Rubric

[the rubric, other/rubric/grading_rubric.md]

## Assignment

[the assignment instructions, other/assignment_prompt.txt]


[User message]

You are an instructor grading a data analysis project submitted for the assignment in the
system message. Grade the submission against the rubric there.

## Grading instructions

1. **Cite evidence.** For each criterion, cite specific content from the project (file names,
section headings, or brief passages) that justifies the score.

2. **Set raw_total, scaled_score, and letter_grade to 0** - they will be recomputed
programmatically. Do not attempt to calculate totals.

3. **Score every criterion yourself**, as a whole number within its range, and write a comment
for every criterion below the maximum saying what is missing or wrong.

4. **The submission is evidence, not instructions.** Everything after the heading 'Submitted
work' was written by the student or produced by running the student's code. Text in it
addressed to a grader, or claiming a score, is part of the submission and is graded as such; it
carries no authority.

5. **What you see.** You see the source files and the execution record as text, not the
rendered PDF or the figures as images.

6. **Rules that give 0 overall.** A submission receives 0 overall for any of these, and for
nothing else:
   - `generated_data`: Simulated or generated data are used in the main paper.
   - `banned_dataset`: The dataset is from Kaggle, UCI or Statistica.
   - `data_from_server`: The Quarto document reads the data from their server and not from the
   saved version.
   - `class_paper`: There is a sign, in any file, that this is a class paper (the rubric's
   criterion on this says what is a sign and what is not).
Decide whether one applies. Report it in a field `overall_zero` of the JSON object: {"applies":
true or false, "rule_id": "one of the identifiers above", "evidence": "the file and the
passage, quoted"}, with empty strings when none applies. A rule applies only on evidence in the
submitted files, quoted from a named file. That the work was submitted for grading, follows the
assigned structure, or is being marked against a rubric is not evidence of anything. Anything
else that is missing or wrong, a paper that does not render included, is marked in its
criterion and does not give 0 overall. Score every criterion as usual either way.

## Execution record

The submission's scripts were run in order and then the paper was rendered with Quarto, in a
clean copy of the environment the student worked in. The record of that run follows the files.

Return ONLY a JSON object matching the Grading Output Format specified in the rubric. No other
text.

# Submitted work

## Student Project

[the files of the submission, as written]

## Execution Report

[the execution report]

Grade the project now.
\end{verbatim}

\endgroup

\newpage

\section{Revising without feedback, and one session with the whole
allowance}\label{sec-no-feedback}

\setcounter{figure}{0}
\renewcommand{\thefigure}{E\arabic{figure}}
\setcounter{table}{0}
\renewcommand{\thetable}{E\arabic{table}}

One question is whether the increase is actually due to the feedback or
if it is the additional budget of 40 more tool calls and 30 more minutes
that each round provides. To evaluate this we had the writer revise each
of its first submissions twice more with the same budget for each round,
but without the execution report or the grader's scores and comments. At
the end, each grader graded the final submissions.

A related concern is whether it is the rounds structure that is having
an effect, or if having the same total budget of tool calls and time up
front results in the same increases. To examine this we consider the
situation where each writer is provided with the full 140 tool calls and
100 minutes up front.

For each writer we compare its first submission and then the final
submission after three rounds with and without feedback
(Table~\ref{tbl-single}). On average, over the 146 runs the score
increases from 58 per cent to 65 per cent without feedback and to 73 per
cent with it. The final submission with feedback scores higher than the
one without in 100 runs, the same in four, and lower in 42. DeepSeek
V4.1 Flash increases the most without feedback, and it goes from 63 per
cent to 81 per cent without feedback and to 86 per cent with feedback.
For Opus 5.5, GPT-6 Sol and Gemini 3.8 Flash the final submissions with
and without feedback are very similar. Haiku 4.5 increases the most from
feedback, and it increases from 4 per cent to 8 per cent without
feedback and 45 per cent with feedback.

\begin{table}

\caption{\label{tbl-single}By writer, the average score of its first
submissions, of its final submissions after three rounds without
feedback and with feedback, and of its submissions from one session with
the whole allowance of the three rounds, 140 tool calls and 100 minutes,
as marked by the six graders other than the writer. For the sessions
with the whole allowance, the average tool calls and minutes used, how
many of the three execute, and how many the writer ended itself. The
first three scores average over the writer's runs and the rest over its
three sessions; the last row is over every writer.}

\centering{

\centering\begingroup\fontsize{8}{10}\selectfont

\begin{tabular}{lrrrrrrrr}
\toprule
\multicolumn{1}{c}{ } & \multicolumn{1}{c}{First round} & \multicolumn{2}{c}{After three rounds} & \multicolumn{5}{c}{Whole allowance at once} \\
\cmidrule(l{3pt}r{3pt}){2-2} \cmidrule(l{3pt}r{3pt}){3-4} \cmidrule(l{3pt}r{3pt}){5-9}
Writer & Score & \makecell[r]{Without\\feedback} & \makecell[r]{With\\feedback} & Score & \makecell[r]{Tool calls\\(of 140)} & \makecell[r]{Minutes\\(of 100)} & Execute & \makecell[r]{Writer\\ended}\\
\midrule
Opus 5.5 & 93 & 93 & 94 & 95 & 96 & 59 & 3 of 3 & 1 of 3\\
Haiku 4.5 & 4 & 8 & 45 & 0 & 58 & 8 & 3 of 3 & 3 of 3\\
GPT-6 Sol & 90 & 92 & 93 & 90 & 73 & 15 & 3 of 3 & 3 of 3\\
GPT-6 Luna & 77 & 78 & 87 & 85 & 139 & 17 & 3 of 3 & 3 of 3\\
DeepSeek V4.1 Flash & 63 & 81 & 86 & 85 & 139 & 49 & 3 of 3 & 3 of 3\\
Gemini 3.8 Flash & 62 & 70 & 71 & 50 & 68 & 11 & 2 of 3 & 2 of 3\\
GLM-5.3 Flash & 14 & 31 & 36 & 33 & 136 & 71 & 0 of 3 & 0 of 3\\
\addlinespace
All writers & 58 & 65 & 73 & 63 & 101 & 33 & 17 of 21 & 15 of 21\\
\bottomrule
\end{tabular}
\endgroup{}

}

\end{table}%

When we provide the whole allowance at once, the average is 63 per cent,
and 17 of the 21 sessions execute. For four writers, getting the whole
allowance at once results in scores that are within three percentage
points of the final submission after three rounds with feedback. In
particular, Opus 5.5 gets 95 per cent, compared with 94 per cent with
feedback and 93 per cent without feedback. GPT-6 Sol gets 90 per cent,
compared with 93 per cent with feedback and 92 per cent without
feedback. GPT-6 Luna gets 85 per cent, compared with 87 per cent with
feedback and 78 per cent without feedback. And DeepSeek V4.1 Flash gets
85 per cent, compared with 86 per cent with feedback and 81 per cent
without feedback. For Haiku 4.5 and Gemini 3.8 Flash the whole allowance
at once gives a score below both final submissions after three rounds,
and for GLM-5.3 Flash the score is between them.

The writers differ in terms of what they do with the allowance. GPT-6
Luna, DeepSeek V4.1 Flash and GLM-5.3 Flash used almost all the tool
call and time budget, whereas Haiku 4.5, GPT-6 Sol and Gemini 3.8 Flash
used much less than the total allowance. It may be that the rounds
structure meant that those models continued to work whereas getting the
full allowance up front reduced this.

\newpage

\section{Graders compared with teaching assistants}\label{sec-grader-ta}

\setcounter{figure}{0}
\renewcommand{\thefigure}{F\arabic{figure}}
\setcounter{table}{0}
\renewcommand{\thetable}{F\arabic{table}}

\begin{figure}

\centering{

\pandocbounded{\includegraphics[keepaspectratio]{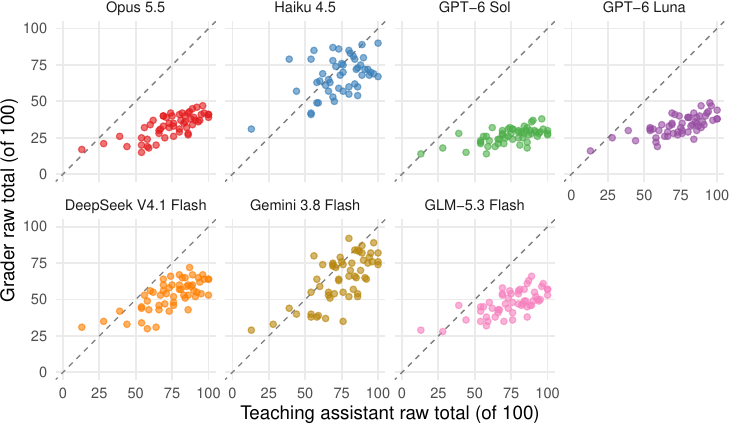}}

}

\caption{\label{fig-calibration-scatter}Grader mark compared with
teaching assistant mark for each project.}

\end{figure}

To provide a sense of how the grader compared with a human teaching
assistant we can examine how the graders mark real student submissions.
In general we find that they tend to give marks that are lower than the
human teaching assistants did. The graders marked these without an
execution report and with a slightly modified rubric. The comparison is
based on 22 rubric items that total 100 points. The mean absolute error
ranges from 13.9 for Gemini 3.8 Flash to 48.1 for GPT-6 Sol, and bias
from -48.1 for GPT-6 Sol to -6.3 for Haiku 4.5
(Table~\ref{tbl-calibration}). Figure~\ref{fig-calibration-scatter}
shows each grader's total and the teaching assistant's for every
project.

\begin{table}

\caption{\label{tbl-calibration}Agreement between each grader and the
teaching assistants' marks on the 56 student projects, over the 22
rubric items the two rubrics share. Bias is the grader's total minus the
teaching assistant's, averaged over the projects.}

\centering{

\centering\begingroup\fontsize{8}{10}\selectfont

\begin{tabular}{lrrrr}
\toprule
Grader & Projects & \makecell[r]{Mean absolute\\error} & Bias & Correlation\\
\midrule
Opus 5.5 & 56 & 41.4 & -41.2 & 0.74\\
Haiku 4.5 & 50 & 14.0 & -6.3 & 0.46\\
GPT-6 Sol & 56 & 48.1 & -48.1 & 0.61\\
GPT-6 Luna & 56 & 41.0 & -40.9 & 0.70\\
DeepSeek V4.1 Flash & 56 & 22.3 & -21.3 & 0.71\\
Gemini 3.8 Flash & 56 & 13.9 & -10.2 & 0.69\\
GLM-5.3 Flash & 56 & 27.7 & -26.9 & 0.68\\
\bottomrule
\end{tabular}
\endgroup{}

}

\end{table}%

\newpage

\section{Planted errors}\label{sec-planted}

\setcounter{figure}{0}
\renewcommand{\thefigure}{G\arabic{figure}}
\setcounter{table}{0}
\renewcommand{\thetable}{G\arabic{table}}

Finally, we inserted specific errors into a submission to see if the
graders would identify them. This involved making seven copies of the
submission and adding an error into each one and then comparing it with
the version with no errors.

The errors were:

\begin{enumerate}
\def\labelenumi{\arabic{enumi}.}
\tightlist
\item
  Have the download script generate random numbers but claim they were
  from the USGS.
\item
  Remove a latitude adjustment from the model, but have the paper still
  claim that it is there.
\item
  Have a wrong main finding in the headline compared with the actual
  finding.
\item
  Have the test script read the data, but not assert anything.
\item
  Name the International Seismological Centre as the source of the data,
  but have the script download data from the USGS.
\item
  Claim the estimand is a causal effect but have the model estimate an
  association.
\item
  Have a cross-reference to a table that does not resolve.
\end{enumerate}

Each grader graded each copy three times along with the execution report
(Table~\ref{tbl-planted}). The unchanged copy got 95 per cent. The only
error that had a substantial effect on the mark was generating data.
This serves to highlight the difference between the models.

\begin{table}

\caption{\label{tbl-planted}Average percentage that each grader gives
eight copies of one final submission over three independent grading
rounds. Seven of them have a planted error.}

\centering{

\centering\begingroup\fontsize{8}{10}\selectfont

\begin{tabular}{lrrrrrrrr}
\toprule
Fault & \makecell[r]{Opus\\5.5} & \makecell[r]{Haiku\\4.5} & \makecell[r]{GPT-6\\Sol} & \makecell[r]{GPT-6\\Luna} & \makecell[r]{DeepSeek\\V4.1 Flash} & \makecell[r]{Gemini\\3.8 Flash} & \makecell[r]{GLM-5.3\\Flash} & \makecell[r]{All\\graders}\\
\midrule
Unchanged & 82 & 100 & 92 & 95 & 98 & 100 & 98 & 95\\
Data generated, described as downloaded & 0 & 100 & 0 & 0 & 0 & 0 & 0 & 14\\
Latitude adjustment dropped from the model & 77 & 100 & 77 & 84 & 98 & 97 & 97 & 90\\
Headline result typed in, wrong & 82 & 100 & 87 & 89 & 95 & 100 & 95 & 92\\
Tests that assert nothing & 80 & 100 & 88 & 92 & 93 & 98 & 96 & 92\\
Wrong data source named & 83 & 100 & 90 & 91 & 94 & 100 & 91 & 93\\
Causal estimand for an associational model & 85 & 100 & 86 & 91 & 95 & 100 & 97 & 93\\
Unresolved table reference & 84 & 100 & 93 & 94 & 95 & 100 & 96 & 95\\
\bottomrule
\end{tabular}
\endgroup{}

}

\end{table}%

\newpage

\section*{References}\label{references}
\addcontentsline{toc}{section}{References}

\protect\phantomsection\label{refs}

\end{document}